\documentclass[11pt]{article}
\usepackage{times}
\usepackage{amsmath}
\usepackage{amssymb}
\usepackage[ansinew]{inputenc}
\usepackage{bm}
\usepackage{float}
\usepackage{url}
\usepackage{algorithm}
\usepackage{algorithmic}
\usepackage{graphicx}
\usepackage{subfigure}
\usepackage[small]{caption}
\usepackage{color}
\usepackage{graphics}
\usepackage{graphicx}
\usepackage{multirow}

\newcommand{\boldK}{\mathbf{K}} 
\newcommand{\eye}{\mathbf{I}}   
\newcommand{\boldH}{\mathbf{H}} 

\newcommand{\dirFigures}{./figures/}

\title{Short-term time series prediction using Hilbert space embeddings of autoregressive processes}
\author{Edgar A. Valencia$^\sharp$, Mauricio A. \'Alvarez$^{\dagger}$\\
{\small $\sharp$ \emph{Department of Mathematics, Universidad Tecnol{\'o}gica de Pereira, Colombia, 
660003.}}\\
{\small $\dagger$ \emph{Faculty of Engineering, Universidad Tecnol{\'o}gica de Pereira, Colombia, 
660003.}}\\
\\}
\date{}

\begin{document}
\maketitle

\begin{abstract}

Linear autoregressive models serve as basic representations of
  discrete time stochastic processes. Different attempts have been
  made to provide non-linear versions of the basic autoregressive
  process, including different versions based on kernel methods.
  Motivated by the powerful framework of Hilbert space embeddings of
  distributions, in this paper we apply this methodology for the
  kernel embedding of an autoregressive process of order $p$. By doing
  so, we provide a non-linear version of an autoregressive process,
  that shows increased performance over the linear model in highly
  complex time series. We use the method proposed for one-step
  ahead forecasting of different time-series, and compare its
  performance against other non-linear methods.
\end{abstract}


\section{Introduction}
\label{sec:intro}

Autoregressive processes are useful probabilistic models for discrete time random processes. The basic idea in an autoregressive process is that the random variable at time $n$, can be described as a linear combination of the $p$ past random variables associated to the process, plus white Gaussian noise. The value of $p$ determines the order of the autoregressive process \cite{Shanmugan}.

Different authors have proposed non-linear extensions of the above model including NARMAX (non-linear autoregressive moving average model with exogenous inputs) \cite{Shumway}, and also including the use of more general non-linear regression methods for extending the classical autoregressive process to non-linear setups. Examples of non-linear regression methods used are neural networks \cite{Nelles}, Gaussian processes \cite{Kocijan}, and kernel-based learning methods \cite{Kallas}.

Within the kernel methods literature, different versions for kernelizing an autoregressive process of order $p$ have been proposed \cite{Kumar, Kallas}. In \cite{Kumar}, the authors propose an AR process built over a feature space. The coefficients of the autoregressive model are estimated by minimizing the quadratic error between the feature map of the input at time $n$, and the prediction given by the linear combination of the last $p$ mapped inputs. Predictions are presented only for finite dimensional feature mappings, for which the inverse mapping from a feature space to the input space is easily computed.  In \cite{Kallas}, the authors also propose an AR process built over a feature space, by this time, the coefficients of the autoregressive model are estimated by using Yule-Walker equations, where the correlations between random variables are replaced by inner products between the feature maps of those random variables. Predictions are obtained by solving a pre-image problem.

Our objective in this paper is to introduce a non-linear version of the autoregressive model of order $p$ based on Hilbert space embeddings of joint probability distributions. 

Hilbert space embeddings are a recent trend in kernel methods that map distributions into infinite-dimensional feature spaces using kernels, such that comparisons and manipulations of these distributions can be performed using standard feature space operations like inner products or projections \cite{Song3}. Hilbert space embeddings have been successfully used as alternatives to traditional parametric probabilistic models like hidden Markov models \cite{Song2} or linear dynamical systems \cite{Song1}. They have also been used as non-parametric alternatives to statistical tests \cite{Smola}.

Motivated by this powerful framework, we develop a kernelized version of an autoregressive model by means of the Yule-Walker algorithm, and instead of computing correlations (as in the classical AR linear model) or inner products (as in \cite{Kallas}), we compute cross-covariance operators for pairs of random variables. For time-series prediction, one additionally needs to solve a pre-image problem \cite{Honeine}, to map from the space of covariance operators to the original input space. We develop an algorithm that uses fixed point iterations for solving the pre-image problem.  The performance of the proposed model is compared against the linear AR model, the kernel method proposed in \cite{Kallas}, neural networks, and Gaussian processes, for one-step ahead forecasting in different time series. 

The paper is organized as follows. In section \ref{sec:hsp}, we briefly review Hilbert space embeddings methods. In section \ref{sec:armodel}, we present the embedding of the AR model using cross-covariance operators, including parameter estimation, and solving the pre-image problem. In section \ref{sec:relatedwork}, we present some related work. In section \ref{sec:experimental:eval} we describe the experimental setup that includes four datasets, and in section \ref{sec:results}, we show the results for one-step ahead prediction over the different datasets. Conclusions appear in section \ref{sec:conclusions}.

\section{Review of Hilbert space embeddings}
\label{sec:hsp}

In this paper, we use upper-case letters to refer to random variables
(for example, $X, Y$), and lower-case letters to refer to particular
values that those random variables can take (for example, $x,
y$). Upper-case bold letters are used to refer to matrices, and lower-case
bold letters are used for vectors.

We briefly review the definitions of a reproducible kernel Hilbert space
(RKHS), Hilbert space embeddings of distributions, and covariance
operators, which are the key for developing Hilbert space
embeddings of autoregressive processes. 

\subsection{Reproducing Kernel Hilbert Space}
A reproducing kernel Hilbert space (RKHS) $\mathcal{H}$ with kernel
$k(x, x')$, for $x,x'\in\mathcal{X}$, is a space of functions $g:\mathcal{X}\rightarrow
\mathbb{R}$ that satisfy the following properties:
\begin{enumerate}
\itemsep 0.01cm
\item For all $x\in \mathcal{X}$, $k(x,\cdot):\mathcal{X}\rightarrow \mathbb{R}$ belongs to $\mathcal{H}$.\\
\item $\langle g(\cdot), k(x,\cdot)\rangle_{\mathcal{H}}=g(x)$ and consequently $\langle k(x,\cdot), k(y,\cdot)\rangle_{\mathcal{H}}=k(x,y).$
\end{enumerate}
An alternative definition for a kernel function, which is usually used when designing
algorithms, is given by $k(x,x')=\langle \phi(x),
\phi(x')\rangle_{\mathcal{H}}$, where
$\phi:\mathcal{X}\rightarrow \mathcal{H}$. 

Kernel methods are widely popular in signal processing and machine
learning, and there are several textbooks where they are described in
detail \cite{Scholkopf:kernels:2002,Shawe:kernels:2004, Bishop}. 

\subsection{Embedding distributions}

Recently, the authors in \cite{Smola} introduced a method for
embedding probability distributions in a RKHS. Let $\mathcal{P}$ be
the space of all probability distributions $\mathbb{P}$ on
$\mathcal{X}$. Let $X$ be a random variable with distribution function
$\mathbb{P}\in \mathcal{P}$. In \cite{Smola}, the authors define the
mapping from a probability distribution $\mathbb{P}\in \mathcal{P}$ to
a RKHS $\mathcal{H}$ using the mean map $\mu_X$ defined as
\begin{equation}
\mu_X(\mathbb{P})=\mathbb{E}_{X}[k(X,\cdot)]=\mathbb{E}_{X}[\phi(X)].\nonumber
\end{equation}
The mean map $\mu_X$ satisfies $\langle \mu_X,
\phi(\cdot)\rangle_{\mathcal{H}}=\mathbb{E}_X[\phi(X)]$. If the kernel $k(x,x')$
used for the embedding is \textit{characteristic}, \footnote{A characteristic kernel is a reproducing kernel for which
  $\mu_X(\mathbb{P})=\mu_Y(\mathbb{Q})\iff \mathbb{P}=\mathbb{Q}$, $\mathbb{P},\mathbb{Q}\in\mathcal{P}$, where $\mathcal{P}$ denotes the set of all Borel probability measures on a topological space $(M,\mathcal{A})$.} then $\mu_X$ is injective.

Given an \textit{i.i.d.} set of observations $\{x^l\}_{l=1}^m$ of the
random variable $X$, an estimator for $\widehat{\mu}_X$ is given as
\begin{equation*}
 \widehat{\mu}_X=\frac{1}{m}\sum^m_{l=1}k(x^l,\cdot).
 \end{equation*}
It can be shown  that $\langle \widehat{\mu}_X, \phi(\cdot)\rangle_{\mathcal{H}}=\frac{1}{m}\sum^m_{l=1}\phi(x^l)$.
The estimator $\widehat{\mu}_X$ converges to $\mu_X$, in the norm of
$\mathcal{H}$, at a rate of $O_p(m^{-1/2})$  (see \cite{Smola} for details). 

\subsection{Cross-covariance operator}

If $\mathcal{H}_1$ and $\mathcal{H}_2$ are RKHS with kernels
$k(\cdot, \cdot)$ and $\ell(\cdot, \cdot)$, and feature maps $\phi$
and $\varphi$, respectively,  the \textit{uncentered cross-covariance
  operator} is defined as \cite{Baker}
\begin{align*}
\mathcal{C}_{XY}=\mathbb{E}_{XY}[\phi(X)\otimes\varphi(Y)],
\end{align*}
where $\otimes$ is the \emph{tensor product}.\footnote{Given $f, h\in \mathcal{H}_1$, and $g\in \mathcal{H}_2$, we define the
tensor product $f\otimes g$ as an operator that maps $h$ from
$\mathcal{H}_1$ to $\mathcal{H}_2$ such that $(f\otimes g)h\rightarrow \langle h, f \rangle_{\mathcal{H}_1} g$.}
The cross-covariance operator $\mathcal{C}_{XY}$ can be seen as an 
element of a \textit{tensor product reproducing kernel Hilbert space} (TP-RKHS),
$\mathcal{H}_1 \otimes \mathcal{H}_2$.

Given two functions $f\in \mathcal{H}_1$ and $g\in \mathcal{H}_2$ then
\begin{align*}\label{2}
\langle f,\mathcal{C}_{XY}g\rangle_{\mathcal{H}_1}&= 
\langle f \otimes g, \mathcal{C}_{XY}\rangle_{\mathcal{H}_1\otimes
                                           \mathcal{H}_2}\\
&=\mathbb{E}_{XY}\left[\langle f\otimes g,
  \phi(X)\otimes\varphi(Y)\rangle_{\mathcal{H}_1\otimes
  \mathcal{H}_2}\right] \\
&=\mathbb{E}_{XY}\left[\langle f,\ \phi(X)\rangle_{\mathcal{H}_1}\langle g,\
  \varphi(Y)\rangle_{\mathcal{H}_2}\right]\\
&=\mathbb{E}_{XY}\left[f(X) g(Y)\right],
\end{align*}
where $\phi(x)=k(x,\cdot)$, $\varphi(y)=l(y,\cdot)$, and
$\mathbb{E}_{XY}\left[f(x) g(y)\right]$ is the covariance matrix (for details
see \cite{Gretton}). 

The operator $\mathcal{C}_{XY}$ allows the embedding of the set of joint distributions $\mathcal{P}(X,Y)$ in the TP-RKHS $\mathcal{H}_1 \otimes \mathcal{H}_2$.

Given an \textit{i.i.d} of set of pairs of observations
$\mathcal{D}_{XY}=\{(x^1,y^1), (x^2,y^2),\cdots,$ $(x^m,y^m)\}$, 
a cross-covariance estimator ${\bf \widehat{C}}_{XY}$ for $\mathcal{C}_{XY}$ is defined as:
\begin{equation}\label{eq:cross:cov:estimator}
\widehat{\mathbf{C}}_{XY}=\frac{1}{m}\sum_{l=1}^m\phi(x^l)\otimes \varphi(y^l)=\frac{1}{m}{\bm{\Phi}\bm{\Upsilon}^{\top}},
 \end{equation}
where
${\bm{\Phi}}=(\phi(x^1),\phi(x^2),\ldots,\phi(x^m))$, and
${\bm{\Upsilon}}=(\varphi(y^1),\varphi(y^2),\ldots,\varphi(y^m))$ are design matrices
\cite{Bishop}.

\section{Hilbert space embedding of an autoregressive process}
\label{sec:armodel}

In this section, we describe how the basic autoregressive model can be
embedded in a TP-RKHS. We then provide an estimation method for the
parameters of the embedded method, by means of the Yule-Walker
equations. Finally, we describe a procedure for solving the pre-image
problem for the kernel embedding of the autoregressive process. We
solve the pre-image problem for forecasting in time-series.

\subsection{Autoregressive models in TP-RKHS}
Let $X_1,X_2, \cdots, X_n$ a stationary discrete time stochastic process.  A $p$-order AR
model is defined by \cite{Shanmugan}
\begin{align}\label{eq:ar:model}
 X_{i} &=\lambda_{1}X_{i-1}+\lambda_{2} X_{i-2}+\cdots+\lambda_{p}X_{i-p}+\epsilon_{i}= \sum^p_{j=1}
 \lambda_{j} X_{i-j}+\epsilon_{i},
\end{align}
for $i=p+1, p+2,\cdots,n$, where
$\lambda_{1},\lambda_{2},\cdots,\lambda_{p}$ are the model parameters,
and $\epsilon_i$ is white noise with
$\mathbb{E}(\epsilon_i)=0$ and
$\operatorname{var}(\epsilon_i)=\sigma^2$.  We use $\bm{\lambda} =
[\lambda_1, \lambda_2,  \ldots, \lambda_p]^{\top}$. 

The Yule-Walker equations are a set of linear of equations used to estimate
the coefficients $\bm{\lambda}$. The basic idea is to define a set of
$p$ linear equations, where the unknowns are the $p$ coefficients in $\bm{\lambda}$. 
Each linear equation in the Yule-Walker system is formed by computing
the covariance between $X_i$, and $X_{i-k}$ according to
\begin{align*}
\langle X_i, X_{i-k} \rangle & = \sum^p_{j=1}
 \lambda_{j} \langle X_{i-j}, X_{i-k} \rangle +  \langle \epsilon_{i}, X_{i-k} \rangle,
\end{align*} 
for $k=1, \ldots, p$. Assuming independence between $\epsilon_{i}$, and
$X_{i-k} $, the set of linear equations reduce to
\begin{align}\label{eq:yw:ar:model}
\langle X_i, X_{i-k} \rangle & = \sum^p_{j=1}
 \lambda_{j} \langle X_{i-j}, X_{i-k} \rangle,
\end{align} 
for $k=1, \ldots, p$. Given a set of observations for the discrete
time random process, and a suitable estimator for the covariance terms like
$\langle X_i, X_{i-k} \rangle$, it is possible to solve the set of
equations for estimating $\bm{\lambda}$.

The authors in \cite{Kallas} propose a non-linear extension of the AR
process in \eqref{eq:yw:ar:model}, by applying a non-linear
transformation $\varphi:\mathcal{X}\rightarrow \mathcal{H}$ to the random variables $X_i$ in the AR
model,   
\begin{align}\label{eq:varphi:ar:model}
 \varphi(X_{i})  &= \sum^p_{j=1} \alpha_{j} \varphi(X_{i-j})+\varphi(\epsilon_{i}).
\end{align}
Notice that we use a set of coefficients $\bm{\lambda}$ for the
autoregressive model in $\mathcal{X}$, and a set of coefficients
$\bm{\alpha}=[\alpha_1, \ldots, \alpha_p]^{\top}$ for the autoregressive model in $\mathcal{H}$. To
estimate the parameters $\bm{\alpha}$ in the transformed space, the
authors follow a procedure similar to the Yule-Walker equations, but
instead of computing covariances between random variables like $X_i$,
and $X_{i-k}$, they compute inner products between $\varphi(X_i)$, and
$\varphi(X_{i-k})$. With the proper independence assumptions, the
Yule-Walker system of equations in then given as \footnote{For ease of
  exposition, we have assumed that the transformed random variables
  $\varphi(X_i)$ have been substracted the mean of the transformed
  variable $\mu_{\varphi} = \mathbb{E}_{X_i}[\varphi(X_i)]$.}
\begin{align}\label{eq:yw:varphi:ar:model}
\langle \varphi(X_i), \varphi(X_{i-k}) \rangle & = \sum^p_{j=1}
 \alpha_{j} \langle \varphi(X_{i-j}), \varphi(X_{i-k}) \rangle,
\end{align} 
for $k=1, \ldots, p$. Inner products like the ones above can be
replaced by kernel functions. This is usually known as the
\emph{kernel trick} \cite{Scholkopf:kernels:2002,Shawe:kernels:2004}.
Given a set of observations for the discrete time random process
$\{x_i\}_{i=1}^m$, the following set of equations can be used to
compute $\bm{\alpha}$,
\begin{align}\label{eq:yw:kernel:ar:model}
k(x_i, x_{i-k})& = \sum^p_{j=1}\alpha_{j} k(x_{i-j}, x_{i-k}),
\end{align} 
for $k=1, \ldots, p$. Since the values for $k(x_i, x_{i-j})$, and
$k(x_{i-j}, x_{i-k})$ are themselves random variables that depend on
the values of the observations in a particular time series, and
assuming that the discrete time random process is stationary, the authors in
\cite{Kallas} propose the following set of equations to get an
estimate for $\bm{\alpha}$
\begin{align}\label{eq:yw:expected:kernel:ar:model}
\mathbb{E}[k(x_i, x_{i-k})] & = \sum^p_{j=1}
 \alpha_{j} \mathbb{E}[k(x_{i-j}, x_{i-k})],
\end{align} 
for $k=1, \ldots, p$. Expectations are estimated over the set of
available samples. 

Our key contribution in this paper is that we embedd the
autoregressive model in a TP-RKHS by mapping joint distributions like $\mathbb{P}(X_i, X_{i-k})$, and $\mathbb{P}(X_{i-j},
X_{i-k})$  to points in $\mathcal{H}_1\otimes
\mathcal{H}_2$. Embeddings are performed by using cross-covariance
operators, instead of inner products.

Let us start with Equation \eqref{eq:varphi:ar:model}. If we apply
a tensor product with $\phi(X_{i-k})$, at both
sides of Equation \eqref{eq:varphi:ar:model}, and take expected
values, we obtain
\begin{align}\label{eq:tprkhs:varphi:ar:model}
\mathbb{E}_{X_i,
  X_{i-k}}[\varphi(X_i)\otimes\phi(X_{i-k})]&=\sum^p_{j=1}\alpha_j\mathbb{E}_{X_{i-j},
                                              X_{i-k}}[\varphi(X_{i-j})\otimes\phi(X_{i-k})]\\\notag
& +\mathbb{E}_{\epsilon_i X_{i-k}}[\varphi(\epsilon_i)\otimes\phi(X_{i-k})],
\end{align}
for $k = 1, \cdots, p$. If we assume that $\phi(X_{i-k})$, and
$\varphi(\epsilon_i)$ are uncorrelated, then the expression above
reduces to
\begin{equation}\label{eq:tprkhs:cov:operators:ar:model}
\mathcal{C}_{X_{i}X_{i-k}}  =\sum^p_{j=1}\alpha_j\mathcal{C}_{X_{i-j}X_{i-k}},
\end{equation}
where $\mathcal{C}_{X_{i}X_{i-k}}$, and $\mathcal{C}_{X_{i-j}X_{i-k}}$
are cross-covariance operators, defined as
\begin{align*}
\mathcal{C}_{X_{i}X_{i-k}} & = \mathbb{E}_{X_i,
  X_{i-k}}[\varphi(X_i)\otimes\phi(X_{i-k})]\\
\mathcal{C}_{X_{i-j}X_{i-k}} & = \mathbb{E}_{X_{i-j},
                                              X_{i-k}}[\varphi(X_{i-j})\otimes\phi(X_{i-k})].
\end{align*}

\subsection{Parameter estimation for autoregressive models in TP-RKHS}\label{sec:param:estimation:tprkhs}

In this section, we provide a method for estimating the parameters
$\bm{\alpha}$ in the autoregressive model in Equation
\eqref{eq:tprkhs:cov:operators:ar:model}.  For this, we use the
estimator for the cross-covariance operators, as in Equation \eqref{eq:cross:cov:estimator}.

Let $\mathcal{D}_{X_{i}X_{i-j}}=\lbrace(x^1_{i}, x^1_{i-j}),(x^2_{i},
x^2_{i-j}),\cdots,(x^m_{i}, x^m_{i-j})\rbrace$, for $j=1,2,\cdots,p$,
be different sets of samples drawn \textit{i.i.d} from the
distributions  $\mathbb{P}(X_{i},X_{i-j})$. 
We denote by $\bm{\Phi}_i$ the design matrix built from the elements
$\{\phi(x_i^l)\}_{l=1}^m$, and $\bm{\Upsilon}_{i-j}$ the design matrix
built from the elements $\{\varphi(x^l_{i-j})\}_{l=1}^m$, 
\begin{align*}
\bm{\Phi}_i &=(\phi(x^1_i),\phi(x^2_i),\cdots,\phi(x^m_i)),\\
\bm{\Upsilon}_{i-j} & =(\varphi(x^1_{i-j}),\varphi(x^2_{i-j}),\ldots,\varphi(x^m_{i-j})).\nonumber
\end{align*}
Estimators for the cross-covariance operators
$\mathcal{C}_{X_{i}X_{i-k}}$ and $\mathcal{C}_{X_{i-j}X_{i-k}}$ are
given as (see Equation \eqref{eq:tprkhs:cov:operators:ar:model} and reference \cite{Song2})
\begin{align}\label{eq:cross:cov:operator:ar:model}
\widehat{\mathbf{C}}_{X_{i}X_{i-k}}&=\frac{1}{m}\sum^m_{l=1}\phi(x_i^l)\otimes\varphi(x_{i-k}^l)=\frac{1}{m}\bm{\Phi}_i\bm{\Upsilon}_{i-k}^{\top}\\
\widehat{\mathbf{C}}_{X_{i-j}X_{i-k}}&=
                                       \frac{1}{m}\sum^m_{l=1}\varphi(x_{i-j}^l)\otimes\varphi(x_{i-k}^l)
                                       =\frac{1}{m}\bm{\Upsilon}_{i-j}\bm{\Upsilon}_{i-k}^{\top}. 
\end{align}
Equation \eqref{eq:tprkhs:cov:operators:ar:model} can now be written
approximately as
\begin{equation}\label{eq:tprkhs:cov:operators:estimation:ar:model}
 \bm{\Phi}_i\bm{\Upsilon}_{i-k}^{\top} =\sum^p_{j=1} \alpha_j \bm{\Upsilon}_{i-j}\bm{\Upsilon}_{i-k}^{\top}.
\end{equation}

We pre-multiply Equation
\eqref{eq:tprkhs:cov:operators:estimation:ar:model} by
${\bm\Upsilon}_{i-k}^{\top}$, and post-multiply by ${\bm\Phi}_i$, obtaining
\begin{equation}\label{eq:pre:pos:multiply}
{\bm\Upsilon}_{i-k}^{\top}{\bm\Phi}_i{\bm\Upsilon}^{\top}_{i-k}{\bm\Phi}_i=\sum^p_{j=1}\alpha_j{\bm\Upsilon}_{i-k}^{\top}{\bm\Upsilon}_{i-j}{\bm\Upsilon}^\top_{i-k}{\bm\Phi}_i.
\end{equation}
Simplifying
\begin{equation}\label{eq:pre:pos:multiply:simplified}
{\bm\Upsilon}_{i-k}^{\top}{\bm\Phi}_i=\sum^p_{j=1}\alpha_j{\bm\Upsilon}_{i-k}^{\top}{\bm\Upsilon}_{i-j}.
\end{equation}
We can write the expression above as
\begin{equation}\label{eq:general:cross:cov:alpha}
{\bf H}_{i-k, i} =\sum^p_{j=1}\alpha_j{\bf K}_{i-k, i-j}, 
\end{equation}
where ${\bf H}_{i-k, i}= {\bm\Upsilon}_{i-k}^{\top}{\bm\Phi}_i$,
${\bf K}_{i-k, i-j}={\bm\Upsilon}_{i-k}^{\top}{\bm\Upsilon}_{i-j}$,
and $k=1,2,\ldots, p$. Notice that the entries for the matrix
${\bf H}_{i-k, i}$ are the inner products
$\{\varphi(x_{i-k}^r)^{\top}\phi(x_i^s)\}_{r=1, s=1}^{m,m}$. These
inner products can be computed using a kernel function
$\{h(x_{i-k}^r,x_i^s) \}_{r=1,s=1}^{m,m}$. Likewise, entries of
${\bf K}_{i-k, i-j}$ are given by inner products
$\{\varphi(x_{i-k}^r)^{\top}\phi(x_{i-j}^s)\}_{r=1, s=1}^{m,m}$, which
again can be computed using a kernel function $\{k(x_{i-k}^r,x_{i-j}^s) \}_{r=1,s=1}^{m,m}$.

Given a time-series dataset and a value for $p$, the values of ${\bf
  H}_{i-k, i}$, and ${\bf K}_{i-k, i}$ depend on the values chosen for
$i$, and $m$. Assuming that the discrete time random process is
stationary, we can get an estimate for $\bm{\alpha}$ using the
following set of equations

\begin{equation}\label{eq:general:expected:cross:cov:alpha}
\mathbb{E}[{\bf H}_{k}] =\sum^p_{j=1}\alpha_j\mathbb{E}[{\bf K}_{k, j}], 
\end{equation}
for $k=1, \ldots, p$. We have suppresed the subindex $i$ from the
equation above to keep the notation uncluttered. As in Equation in \eqref{eq:yw:expected:kernel:ar:model},
expectations can be estimated over the set of
available samples. 

We can use the system of equations
in \eqref{eq:general:expected:cross:cov:alpha}  to estimate the
parameters $\bm{\alpha}$. The system of equations is given as

\begin{align}\label{eq:ref:explicit:system:equations}
\begin{bmatrix}
\boldH_1\\
\boldH_2\\
\vdots\\
\boldH_p\\
\end{bmatrix} & =
\begin{bmatrix}
\boldK_{1,1} & \boldK_{1,2} & \cdots & \boldK_{1,p}\\
\boldK_{2,1} & \boldK_{2,2} & \cdots & \boldK_{2,p}\\
\vdots          & \vdots         & \vdots & \vdots \\
\boldK_{p,1} & \boldK_{p,2} & \cdots & \boldK_{p,p}\\
\end{bmatrix}
\begin{bmatrix}
\alpha_1 \eye\\
\alpha_2 \eye\\
\vdots\\
\alpha_p \eye\\
\end{bmatrix}
\end{align}
where $\eye$ is the identity matrix of dimension $m$. We can find an estimator for $\bm{\alpha}$ by solving
\begin{align}\label{eq:min:alpha}
\widehat{\bm{\alpha}} & = \arg\min_{\bm{\alpha}}\left\Vert {\boldH - \boldK{\bm\alpha}_m}\right\Vert_2^2,
\end{align}
where $\boldH\in\mathbb{R}^{mp\times m}$ is a block-wise matrix with
blocks given by $\{\boldH_k\}_{k=1}^p$; $\boldK\in\mathbb{R}^{mp\times
mp}$ is a block-wise matrix with blocks given by
$\{\boldK_{k,j}\}_{k=1, j=1}^{p,p}$; and $\bm{\alpha}_m\in\mathbb{R}^{mp\times m}$ is also a
block-wise matrix with blocks given as $\{\alpha_k\eye\}_{k=1}^p$. For
convenience, we also define $\widehat{\boldK}_i\in\mathbb{R}^{mp\times
m}$ as a block-wise matrix taken from $\boldK$, with blocks given by
$\{\boldK_{k, i}\}_{k=1}^p$. 

It can be shown that the optimization problem in \eqref{eq:min:alpha} can be cast into a least-squares
problem as
\begin{align}\label{eq:ls:alpha}
\widehat{\bm{\alpha}} & =\arg\min_{\bm{\alpha}}\left\Vert\mathbf{A} \bm{\alpha}  - \mathbf{b}\right\Vert_2^2,
\end{align}
where $\mathbf{A}\in\mathbb{R}^{p\times p}$ with entries
$\{\operatorname{tr}(\widehat{\boldK}_i^{\top}\widehat{\boldK}_j)\}_{i=1,
j=1}^{p,p}$, and $\mathbf{b}\in\mathbb{R}^{p\times 1}$ with entries $\{\operatorname{tr}(\boldH^{\top}\widehat{\boldK}_i)\}_{i=1}^p$.

\subsection{Solving the pre-image problem for forecasting in a time-series}

We want to use the method above for forecasting a new value $x_i^*$
using $\bm{\alpha}$, and the $p$ previous values of the time
series. For now on, our method allows us to make predictions in the
feature space by means of 
\begin{align}\label{eq:init:pimage}
\tau_i^* & = \sum_{j=1}^p\alpha_j \varphi(x_{i-j}),
\end{align}
where the values for $\{\alpha_j\}_{j=1}^p$ have been estimated as
explained in section \ref{sec:param:estimation:tprkhs}. We would like
to map back the value of $\tau_i^*$ to the input space, to get
the predicted $x_i^*$. In the kernel literature this problem is known
as the \textit{pre-image problem} \cite{Honeine}, and it is an
ill-posed problem due to the higher dimensionality of the feature
space, meaning that the transformed point $\tau_i^*$ may not
have a corresponding $x_i^*$ such that  $\varphi(x_i^*)=\tau_i^*$. 

We apply a tensor product to both sides of expression
\eqref{eq:init:pimage}, leading to 
\begin{align}\label{eq:init:pimage:cov:op}
\tau_i^*\otimes \phi(x_i^*) & = \sum_{j=1}^p\alpha_j
                                 \varphi(x_{i-j}) \otimes \phi(x_i^*).
\end{align}
In order to get an estimate for $x_i^*$, we can solve the following
minimization problem in $\mathcal{H}_1\otimes \mathcal{H}_2$
\begin{align*}
x^*_i & = \arg\min_{x} f(x) = \arg\min_{x}
        \left\Vert \sum^p_{j=1}\alpha_j\varphi(x_{i-j})\otimes\phi(x)-
        \varphi(x)\otimes\phi(x)\right\Vert^2_{\mathcal{H}_1\otimes
        \mathcal{H}_2},
\end{align*}
where we have defined 
\begin{align}\label{eq:fx:norm}
f(x) =\left\Vert 
  \sum^p_{j=1}\alpha_j\varphi(x_{i-j})\otimes\phi(x)- \varphi(x)\otimes\phi(x)\right\Vert^2_{\mathcal{H}_1\otimes
  \mathcal{H}_2}.
\end{align}
Expression for $f(x)$ can also be written as 
\begin{align}\label{eq:fx:innerprod}
f(x)&=\left\langle
      \sum^p_{j=1}\alpha_j\varphi(x_{i-j})\otimes\phi(x),\sum^p_{k=1}\alpha_k\varphi(x_{i-k})\otimes\phi(x)
      \right\rangle_{\mathcal{H}_1\otimes \mathcal{H}_2}\nonumber\\
&-2\left\langle \sum^p_{j=1}\alpha_j\varphi(x_{i-j})\otimes\phi(x), 
   \varphi(x)\otimes\phi(x) \right\rangle_{\mathcal{H}_1\otimes
  \mathcal{H}_2}\nonumber\\
&+\left\langle \varphi(x)\otimes\phi(x), \varphi(x)\otimes\phi(x)
      \right\rangle_{\mathcal{H}_1\otimes \mathcal{H}_2}.
\end{align}
By using the property $\left\langle u\otimes v, a\otimes
  b\right\rangle_{\mathcal{H}_1\otimes \mathcal{H}_2}= \left\langle
  u\otimes a\right\rangle_{\mathcal{H}_1} \left\langle v\otimes
  b\right\rangle_{\mathcal{H}_2}$, we get 
\begin{align}\label{eq:fx:innerprod:property}
f(x)&=\left\langle
      \sum^p_{j=1}\alpha_j\varphi(x_{i-j}),\sum^p_{k=1}\alpha_k\varphi(x_{i-k})
      \right\rangle_{\mathcal{H}_1} \left\langle \phi(x), \phi(x)
      \right\rangle_{\mathcal{H}_2}\nonumber\\
&-2\left\langle \sum^p_{j=1}\alpha_j\varphi(x_{i-j}), 
   \varphi(x)\right\rangle_{\mathcal{H}_1}\left\langle \phi(x), \phi(x)
      \right\rangle_{\mathcal{H}_2}\nonumber\\
&+\left\langle \varphi(x), \varphi(x)\right\rangle_{\mathcal{H}_1}\left\langle \phi(x), \phi(x)\right\rangle_{\mathcal{H}_2}.
\end{align}
Noticing that
$C = \left\langle\sum^p_{j=1}\alpha_j\varphi(x_{i-j}),\sum^p_{k=1}\alpha_k\varphi(x_{i-k})\right\rangle_{\mathcal{H}_1}$
is a constant (it does not depend on $x$), and using kernels $k(x, x')$ of the form $g(\Vert x-
x'\Vert^2)$, we can simplify expression
\eqref{eq:fx:innerprod:property} as follows
\begin{align}\label{eq:fx:innerprod:property:simplified}
f(x)&=Cg(0) - 2 g(0)\sum^p_{j=1}\alpha_jk(x_{i-j}, x) +g^2(0).
\end{align}
Taking the derivative with respect to $x$, we get
\begin{align}\label{eq:fx:derivative}
\frac{d f(x)}{dx}&= - 2 g(0)\sum^p_{j=1}\alpha_j \frac{d k(x_{i-j}, x)}{dx}.
\end{align}
If we use an squared exponential (SE) kernel or a
  radial basis function (RBF) kernel
 \begin{align}\label{eq:se:kernel}
   k(\mathbf{x}, \mathbf{x}') & = \exp\left(-\frac{\|\mathbf{x}
                                -\mathbf{x}'\|^2}{2\ell^2 }\right),
  \end{align} 
  where $\ell^2$ is known as the bandwidth, the expression
  \eqref{eq:fx:derivative} follows as
\begin{align}\label{eq:fx:derivative:se}
\frac{d f(x)}{dx}&= - \frac{2 g(0)}{\ell^2}\sum^p_{j=1}\alpha_j
                   k(x_{i-j}, x) (x_{i-j} - x).
\end{align}
Equating to zero, and solving for $x$, we get the following fixed-point equation 
\begin{align}\label{eq:fix:point}
x_i^* &= \frac{\sum^{p}_{j=1}\alpha_j k(x_{i-j},x^{*}_{i}) x_{i-j}}
{\sum^p_{k=1}\alpha_k k(x_{i-k},x^{*}_{i})}.
\end{align}
\section{Related work}
\label{sec:relatedwork}

As we mentioned in the introduction, the authors in \cite{Kumar}, and \cite{Kallas} introduced a kernelized version of an autoregressive process based on the kernel trick idea \cite{Scholkopf:kernels:2002, Shawe:kernels:2004}. In particular, the autoregressive model is built in a feature space, and the parameters of the model are estimated in two different ways, either by minimizing a quadratic error \cite{Kumar}, or by means of the Yule-Walker algorithm \cite{Kallas}. In \cite{Kumar}, the pre-image problem, this is, the problem of inverse transforming a point in the feature space, to the input space, is only solved for finite-dimensional feature spaces for which the inverse transformation can be readily be computed. In \cite{Kallas}, the pre-image problem is solved by using a fixed-point algorithm similar to equation \eqref{eq:fix:point}. When assuming a stationary kernel, this is $k(x,x') = k(x-x')$, the method in \cite{Kallas} turns out to be a particular example of the system in equation \eqref{eq:general:cross:cov:alpha} for any particular values of $r$, and $s$ in the kernel matrices ${\bf H}_{i-k, i} $, and ${\bf K}_{i-k, i-j}$. 

Expression \eqref{eq:cross:cov:operator:ar:model} follows closely Equation (5.15) in \cite{Bosq:book:2000}. The expression in \cite{Bosq:book:2000} is obtained as the Yule-Walker equations for a so called \emph{autoregressive Hilbertian process of order $p$, ARH(p)}, that corresponds to an autoregressive process defined in a Hilbert space.  In \cite{Bosq:book:2000}, the $\{\alpha_j\}_{j=1}^p$are  bounded linear operators, in contrast to equation  \eqref{eq:cross:cov:operator:ar:model}, where they correspond to scalar values. Estimation of $\alpha_j$, and prediction are different though. The estimation for the bounded linear operators $\{\alpha_j\}_{j=1}^p$ is obtained by projecting the observations in a Hilbert space of finite dimension. Predictions are performed directly by applying the estimated operators over the input data. 

In \cite{Lampert:timeVaryingProb:2015}, the author use kernel mean embeddings to provide one step ahead distribution prediction. In particular, distributions at any time $t$ are represented by kernel mean maps. A mean map at time $t+1$ can be obtained as a mean map at time $t$, linearly transformed by a bounded linear operator. In fact, this corresponds to a ARH(1), where the functions in $\mathcal{H}$ correspond to kernel mean embeddings. The distribution at time $t+1$ in the input space is approximated by a weighted sum of historic input samples. The weigths in the approximation are computed from particular kernel expressions \cite{Lampert:timeVaryingProb:2015}. Our method considers models of order $p$, and our predictions are point estimates in contrast to  \cite{Lampert:timeVaryingProb:2015}. Also, we use embeddings of joint probability distributions, $\mathbb{P}(X_{i}, X_{i-k})$ instead of embeddings of marginal distributions, $\mathbb{P}(X_i)$ (mean maps).

\section{Experimental evaluation}
\label{sec:experimental:eval}

In this section, we provide details for the experimental evaluation performed in this paper. We describe tha datasets we use, and the procedure that we follow for validating the results.

\begin{figure}[ht!]
\centering
\subfigure[Earthrot]{ \label{fig:dataset:earthrot}
\resizebox{0.47\textwidth}{!}{\includegraphics{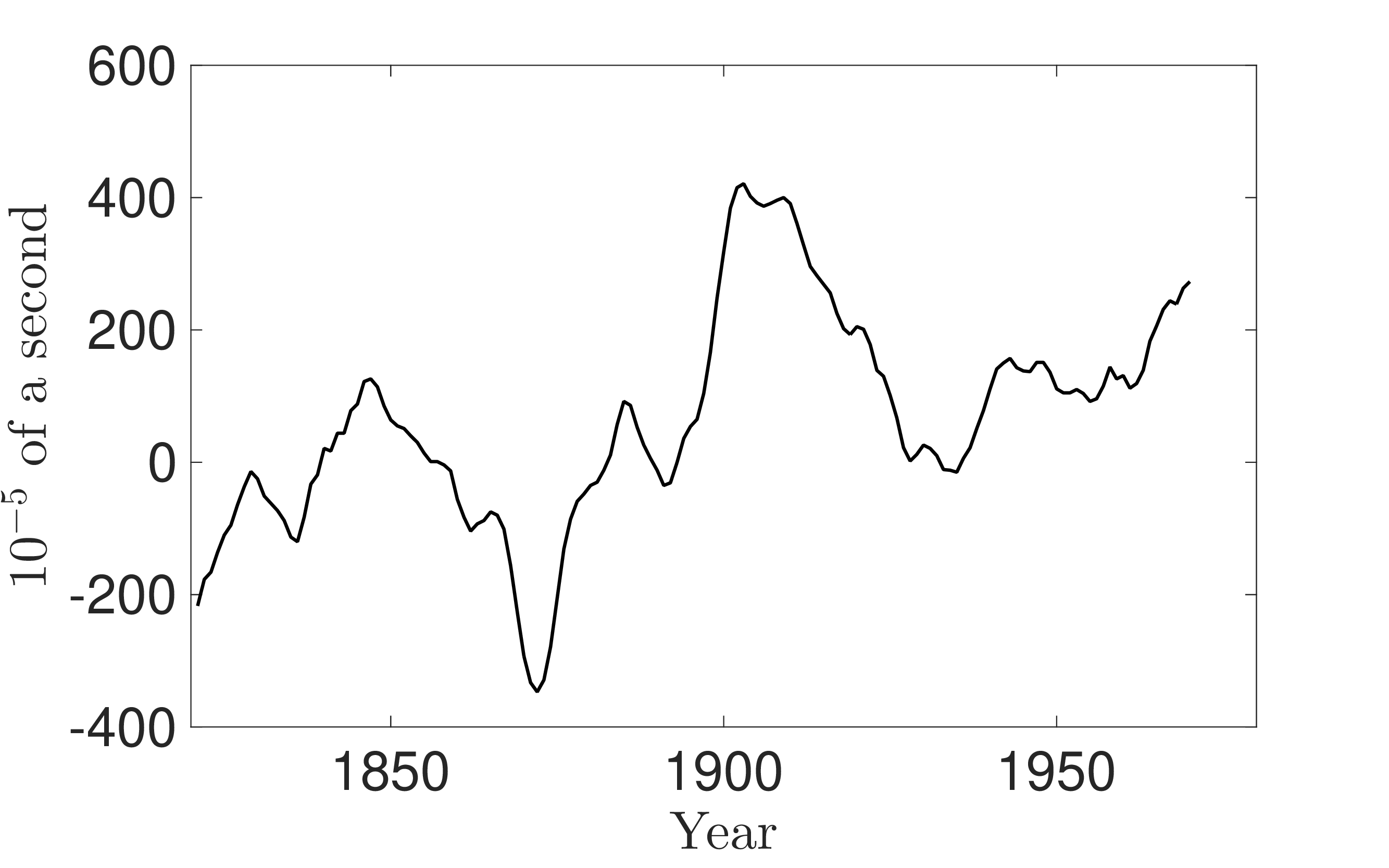}}}
\subfigure[CO$_2$]{ \label{fig:dataset:co2}
\resizebox{0.47\textwidth}{!}{\includegraphics{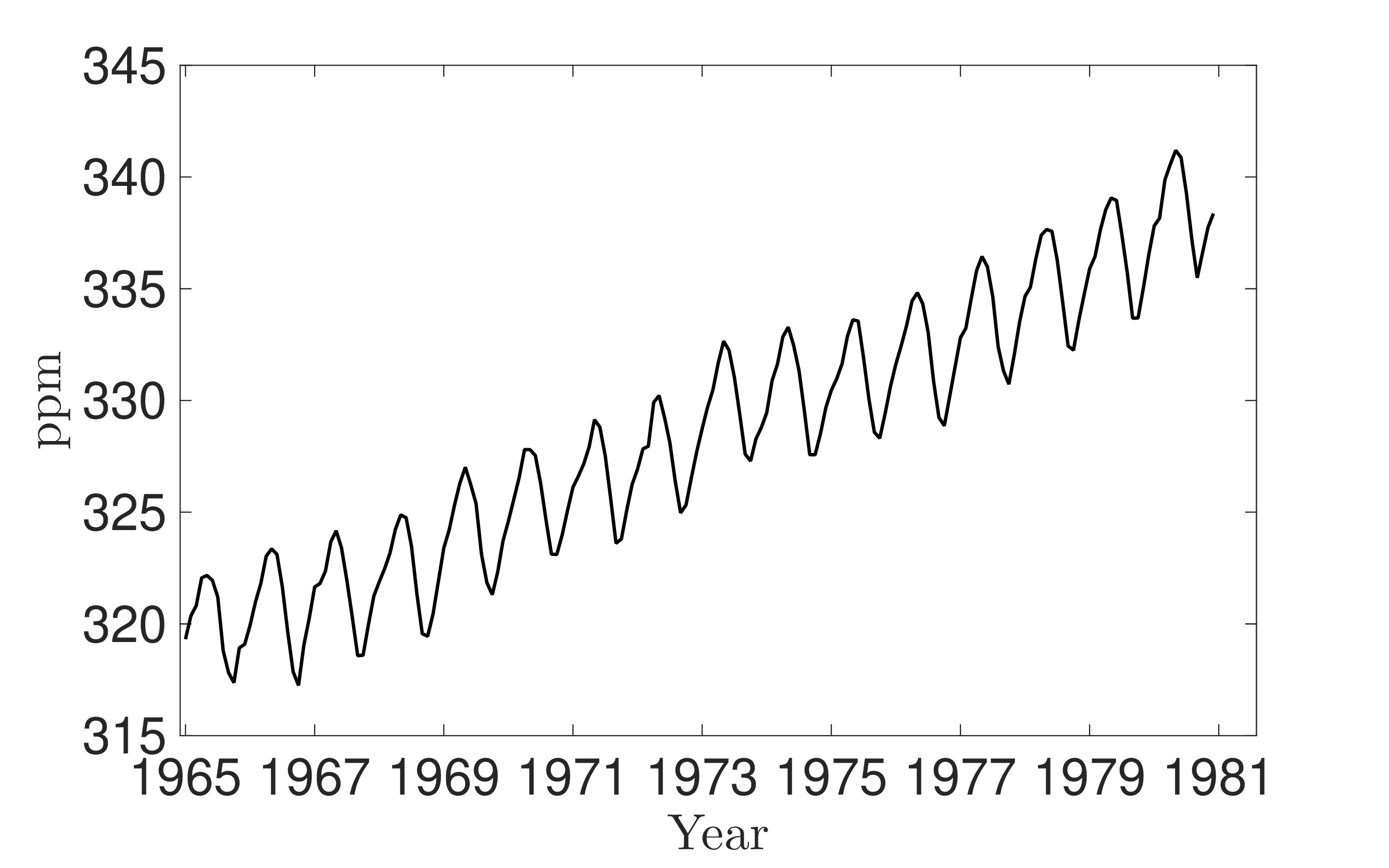}}}
\subfigure[MG$_{30}$]{\label{fig:dataset:mg30}
\resizebox{0.47\textwidth}{!}{\includegraphics{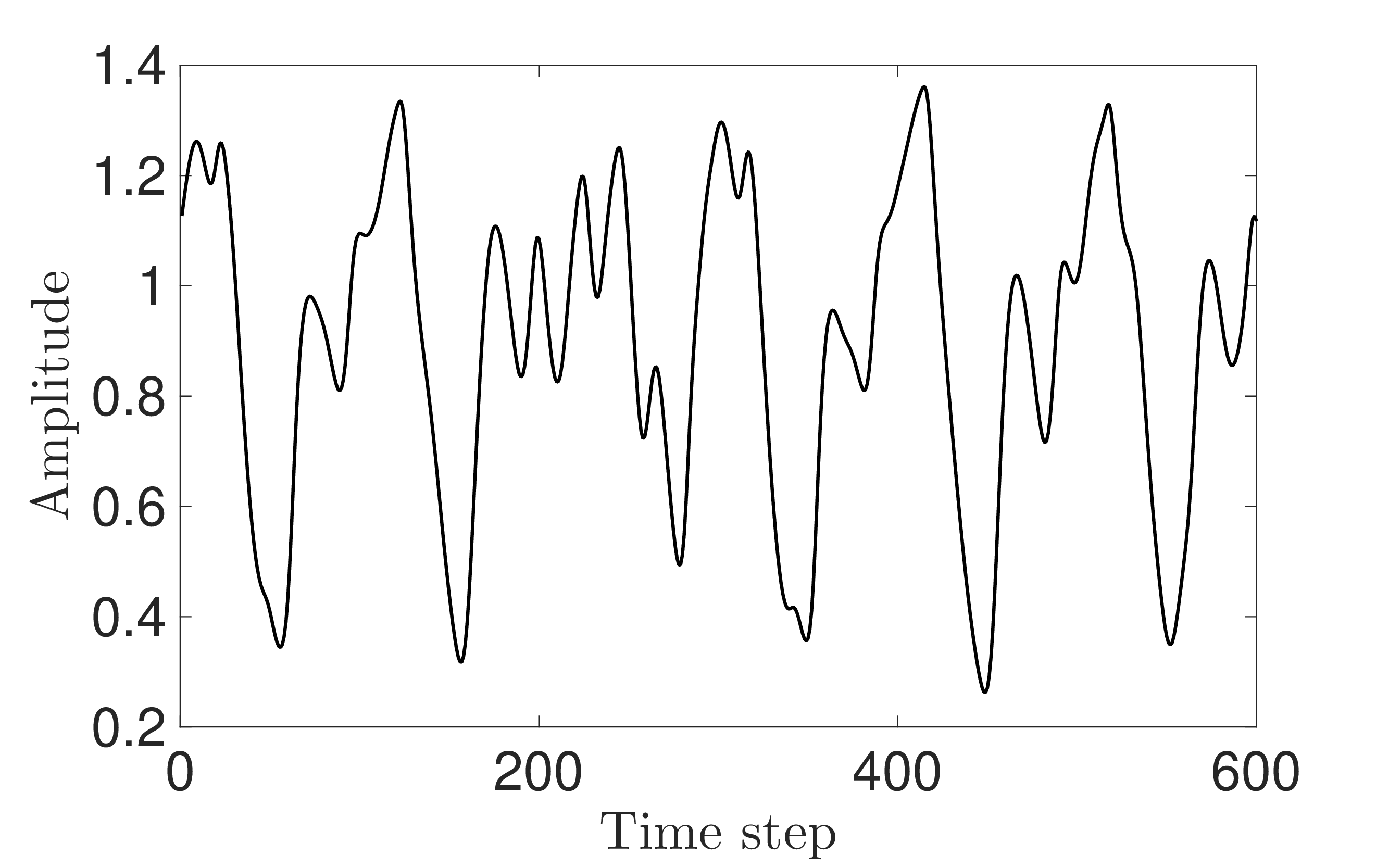}}}
\subfigure[Lorenz]{\label{fig:dataset:lorenz}
\resizebox{0.47\textwidth}{!}{\includegraphics{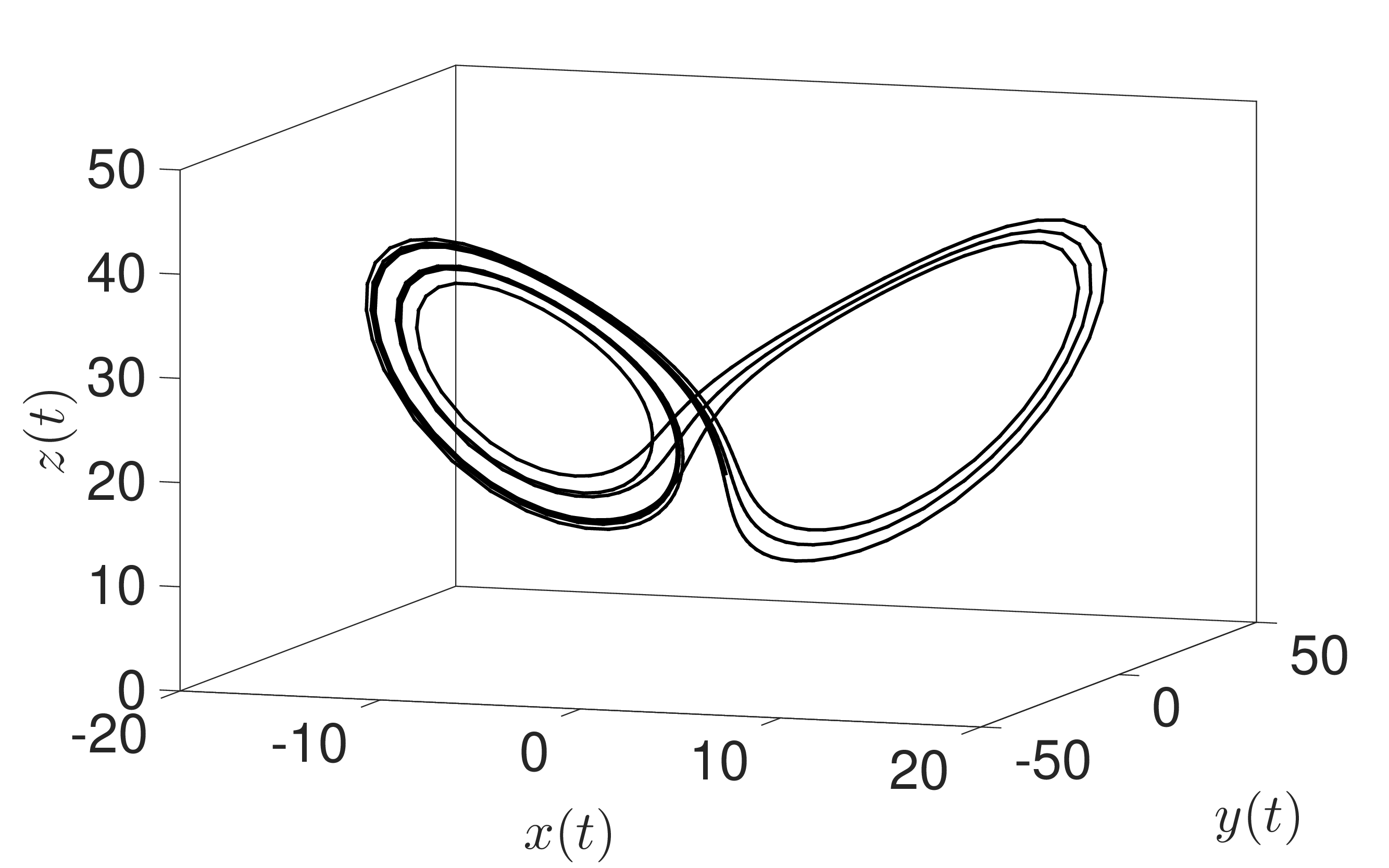}}}
\caption{The four time-series used in this paper to compare the
  performance of the method proposed.}  
\label{fig:datasets}
\end{figure}

\subsection{Datasets}
We use four time-series to evaluate the performance of the different methods. The first two datasets belong to the Time Series Data Library (TSDL), and can be found in \cite{TSDL}. The last two datasets were generated by the authors.

\begin{itemize}
\item[--] \textit{Earthrot}. With the name \textit{Earthrot}, we refer to the Annual changes in the earths rotation, day length (sec*10**-5) 1821-1970 dataset, available at \cite{TSDL}. Units are in $10^{-5}$ of a second. The time-series contains 150 samples. We use the first 130 samples for the experiments.
\item[--] \textit{CO$_2$}. We use the dataset CO2 (ppm) mauna loa, 1965-1980 from the TSDL, which corresponds to monthly measures of CO$_2$ in parts per million from the Mauna Loa observatory. The time-series exhibit a periodic, and approximately linear behavior. The dataset contains 192 samples. For the experiments we use the first 150 samples.
\item[--] \textit{MG$_{30}$}. The time-series MG$_{30}$ refers to the time-series obtained from 
the Mackey-Glass non-linear time delay differential equation given as 
\begin{equation*}
\frac{dx(t)}{dt} = -0.1 x(t) + \frac{0.2x(t-\tau)}{1+x(t-\tau)},
\end{equation*}
with $\tau=30$ \cite{Kallas}. This time series exhibits chaotic dynamics. We generate a time-series of length 600. For the experiments, we use the first 400 samples.
\item[--] \textit{Lorenz}. The Lorenz attractor refers to a set of three coupled ordinary differential equations given as 
\begin{align*}
\frac{dx(t)}{dt} & = -ax+ay\\
\frac{dy(t)}{dt} & = -xz+rx-y\\
\frac{dz(t)}{dt} & = xy-bz,
\end{align*}
where $a, r$, and $b$ are constants. For certain values $a, r$, and $b$, the system exhibits chaotic behavior. We set values for the parameters as $a=10$, $r=28$, and $b=8/3$. For these values, the three-dimensional multi-variate time-series $(x(t), y(t), z(t))$ displays chaotic dynamics. We generate 500 samples per output dimension, and use the first 400 samples for the experiments. We perform prediction over the three time-series $x(t)$, $y(t)$, and $z(t)$, treating them as independent from each other.  
\end{itemize}

Figure \ref{fig:datasets} shows the four datasets described above, and used for testing the methods.


\subsection{Validation}

The validation of the method proposed in this paper is done by
performing one-step ahead prediction over each of the time series
described above. For performing one step-ahead prediction, we use
sliding frames of $w+1$ samples, where the first $w$ samples are used for
training, and the additional last sample is used for validation. The
sliding frames are organized consecutively, with an overlap of $w$ samples. The
training data is used for setting the parameters of each of the models
used for comparison, including the order of the autoregressive
model. For the order of the model, we evaluate values of $p$ from one
to five. We compute the mean-squared error over the validating
samples. We next describe the particular setup used for training in each of the
models used in the experiments. 

\begin{itemize}
\item[--] \textit{Linear AR model (LAR).} The coefficients
  $\bm{\lambda}$ for the linear AR model are estimated using the
  Yule-Waker equations. Within each frame of length $w$, we again use
  a sliding window of size $w/2 + 1$, where the first $w/2$ samples are used to
  compute $\bm{\lambda}$, and the last sample is used for performing
  one-step ahead prediction for different values of $p$. The sliding windows
  are organized consecutively with an overlap of $w/2$ samples. The
  results of the one-ahead step prediction withing the frame of length
  $w$, are used to select the value of $p$, which is selected as the
  value that ocurred more frequently offering the best prediction
  performance. Once the value for $p$ has been selected, we compute
  again the values for $\bm{\lambda}$ using all the datapoints within
  $w$, and used this new $\bm{\lambda}$ for performing one-ahead step
  prediction over the time step $w+1$.
  
\item[--] \textit{Kernel autoregressive model (KAM).} We implement the
  method proposed in \cite{Kallas}. To compute the expectations, we
  use sample means of the quantites of interest. For the kernel
  function, we use an SE kernel as in expression
  \eqref{eq:se:kernel}. The pre-image problem is solved as explained
  in \cite{Kallas}, which has the same fixed-point solution as in
  expression \eqref{eq:fix:point}. The values for $\ell$, and $p$ are
  chosen as follows: within the frame of length $w$, we generate
  sliding frames of size $w/2+1$.  The sliding frames
  are organized consecutively with an overlap of $w/2$ samples.  The first $w/2$ data points are used
  for estimating the values for $\bm{\alpha}$ by solving the system of
  equations in expression \eqref{eq:yw:expected:kernel:ar:model}. We
  then use the data point at time step $w/2+1$ for selecting the best
  value for $\ell$, and $p$, as the ones that on average, within the
  window of length $w$, yield the lowest error. We use a grid of
  values for $\ell$ by taking a grid of percentages, $\ell_p$, of the
  median of the training data within the frame of size $w/2$. The
  percentages that we consider are 0.01, 0.01, 0.5, 1, 2, or 5 of the
  median of the training data within the frame of length $w/2$. Once
  we select the value for $\ell_p$, we compute a new value for $\ell$
  as the percentage $\ell_p$ of the median of the training data within
  the frame of size $w$. Having chosen $\ell$, and $p$, we use all the
  training data of the frame of size $w$ for finding a new set of
  coefficients $\bm{\alpha}$, and finally, provide a forecasting at
  time step $w+1$ by solving again a pre-image problem.

\item[--] \textit{Kernel embedding method (KEM).} We implement the method
  described in section \ref{sec:armodel}. We also use an SE
  kernel. The values for $\ell$, and $p$ used for one-step ahead
  forecasting for time step $w+1$ are computed as follows: within the
  frame of length $w$, we use an sliding window of length $w/2+1$. The sliding windows
  are set up consecutively with an overlap of $w/2$ samples. The
  first $w/2$ data points are used for estimating the coefficients
  $\bm{\alpha}$ by solving equation \eqref{eq:ls:alpha}. For selecting
  $\ell$ and $p$, we follow a similar procedure to the one used for
  the kernel autoregressive model above: the sliding data point at
  time step $w/2 +1$ is used to choose the values for $\ell$, and $p$,
  that on average lead to the lowest prediction errors. Prediction at
  time step $w/2+1$ is performed by solving the pre-image problem 
  in expression \eqref{eq:fix:point}.  In fact, we used percentages of
  the median of the training data, $\ell_p$, in the windows of length
  $w/2$, in order to test different values for $\ell$. The percentages
  that we used were 0.01, 0.01, 0.5, 1, 2, and 5. Once the best
  percentage of the median of the training data for $\ell$, and the
  best value of the order of the model $p$ have been chosen, we
  compute again the value for $\ell$ using the best value for $\ell_p$
  and the training data in the whole frame of size $w$. We again
  compute $\bm{\alpha}$ using the training data in frame $w$, and
  forecast one-step ahead for the time setp $w+1$ solving the
  pre-image problem in expression \eqref{eq:fix:point}.

\item[--] \textit{Gaussian processes (GP).} We follow the model proposed in
  \cite{Kocijan}, in which the random variable of the process at time $X_n$ can be
  described using
\begin{align}\label{eq:ml:equations}
X_n & = f(X_{n-1}, \ldots, X_{n-p}) + \epsilon,
\end{align}
where $\epsilon\sim \mathcal{N}(0, \sigma^2)$, and $f(\mathbf{x})$ is
assumed to follow a Gaussian process prior
$f\sim \mathcal{GP}(0, k(\mathbf{x}, \mathbf{x})))$, with covariance
function $k(\mathbf{x}, \mathbf{x})$. For the covariance function, we
use a SE kernel as in equation \eqref{eq:se:kernel}. The parameter of
the covariance function $\ell$, and the parameter $\sigma$ for the
likelihood model, are estimated by maximizing the log marginal
likelihood using a scaled conjugate gradient procedure. We use the
GPmat Toolbox\footnote{Available at \url{https://github.com/SheffieldML/GPmat}} for all Gaussian processes
related routines. For selecting the value of $p$, we use a sliding
frame of length $w/2 + 1$, within the frame of length $w$. The sliding
windows of length $w/2 + 1$  are established as in the other methods. A number of
$w/2$ data points are used for learning the hyperparameters of the
Gaussian process, and the data point at time step $w/2+1$ is used for
cross-validating the value for $p$. The value for $p$ is chosen as the one
that on average (within the frame of size $w$) leads to the lowest
errors. Once the value for $p$ has been chosen, we use again the $w$ samples
for training a new GP. This new fitted GP is used for forecasting the
data point at time step $w+1$. 

\item[--] \textit{Neural networks (NN).} We use a neural network with one
  hidden layer for learning a similar mapping as in equation
  \eqref{eq:ml:equations}. For choosing the number of neurons $n_h$ of
  the hidden layer, and the value for $p$, we use a similar procedure
  as for the methods above: within the frame of length $w$, we generate
  sliding windows of length $w/2+1$, in a similar way as they were
  slid in the other approaches. The $w/2$ first datapoints are
  used for fitting the weights of the neural network, and the
  data point at time step $w/2+1$ is used for choosing the value for
  $n_h$, and the value for $p$. These values are chosen as the ones
  that on average, within the frame of length $w$, lead to the lowest
  error. We allow the number of neurons in the hidden layer to be any
  of the following values: 5, 10, 15, 20, 25, or 30. For the the
  neural networks routines, we use the Neural Networks toolbox for
  MATLAB, with all the default settings, except for the number of
  neurons in the hidden layer.
\end{itemize}
 
\section{Results}
\label{sec:results}

We compare the performance of the different methods for short-term
prediction over each of the time-series described in  section
\ref{sec:experimental:eval}. Figures \ref{fig:dataset:earthrot},
\ref{fig:dataset:co2}, \ref{fig:dataset:mg30}, and
\ref{fig:dataset:lorenz} show the performance of the classical
linear autoregressive model, the kernel autoregressive model, and the
kernel embeddings of autoregressive model over the four time series
described in Section \ref{sec:experimental:eval}. The mean squared
error (MSE) for one step ahead prediction is shown as the title in each figure.

\begin{figure}[ht!]
\centering
\subfigure[Earthrot using linear AR]{ \label{fig:pred:earthrot:linear}
\resizebox{0.47\textwidth}{!}{\includegraphics{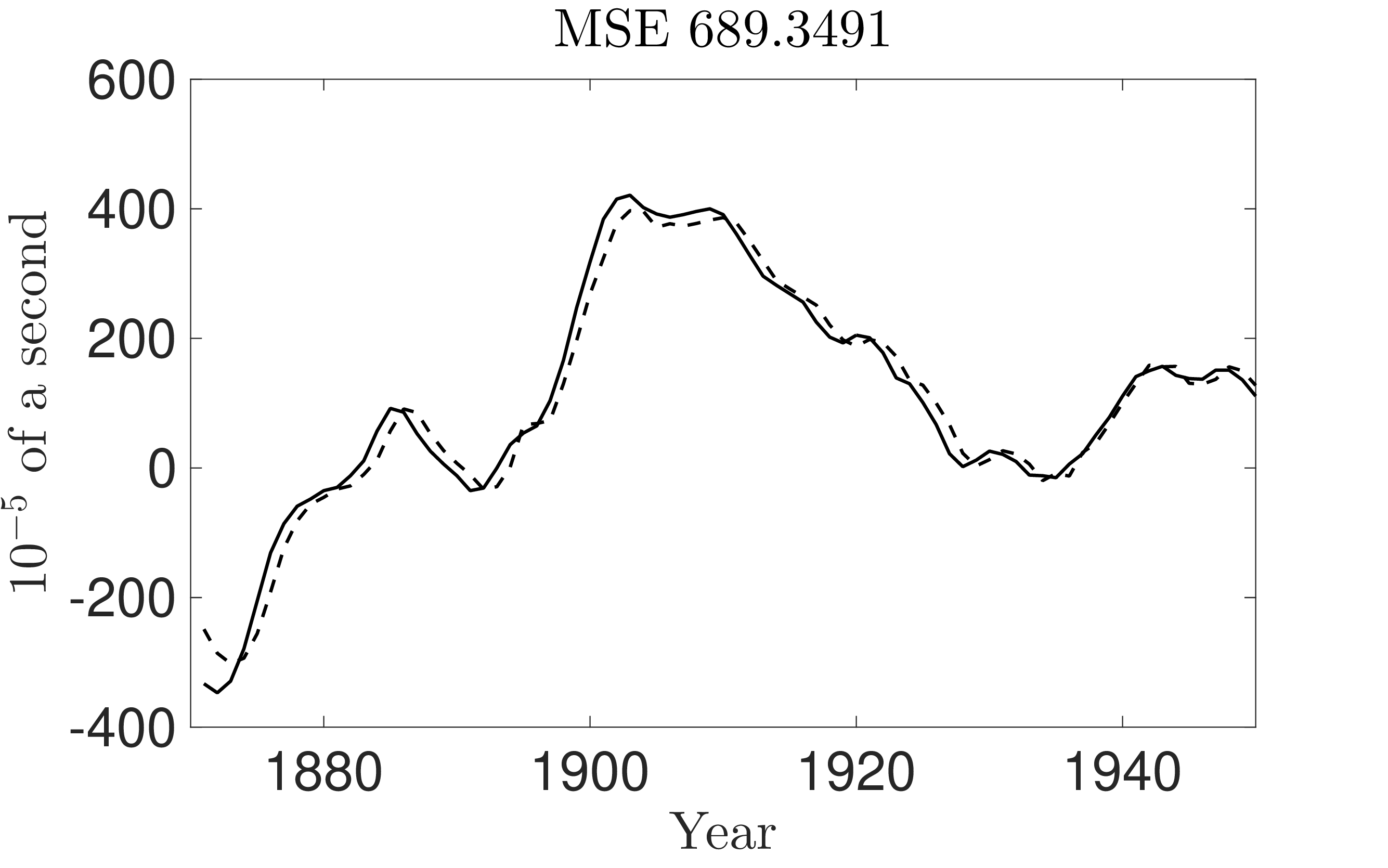}}}
\subfigure[Earthrot using KAM]{\label{fig:pred:earthrot:kallas}
\resizebox{0.47\textwidth}{!}{\includegraphics{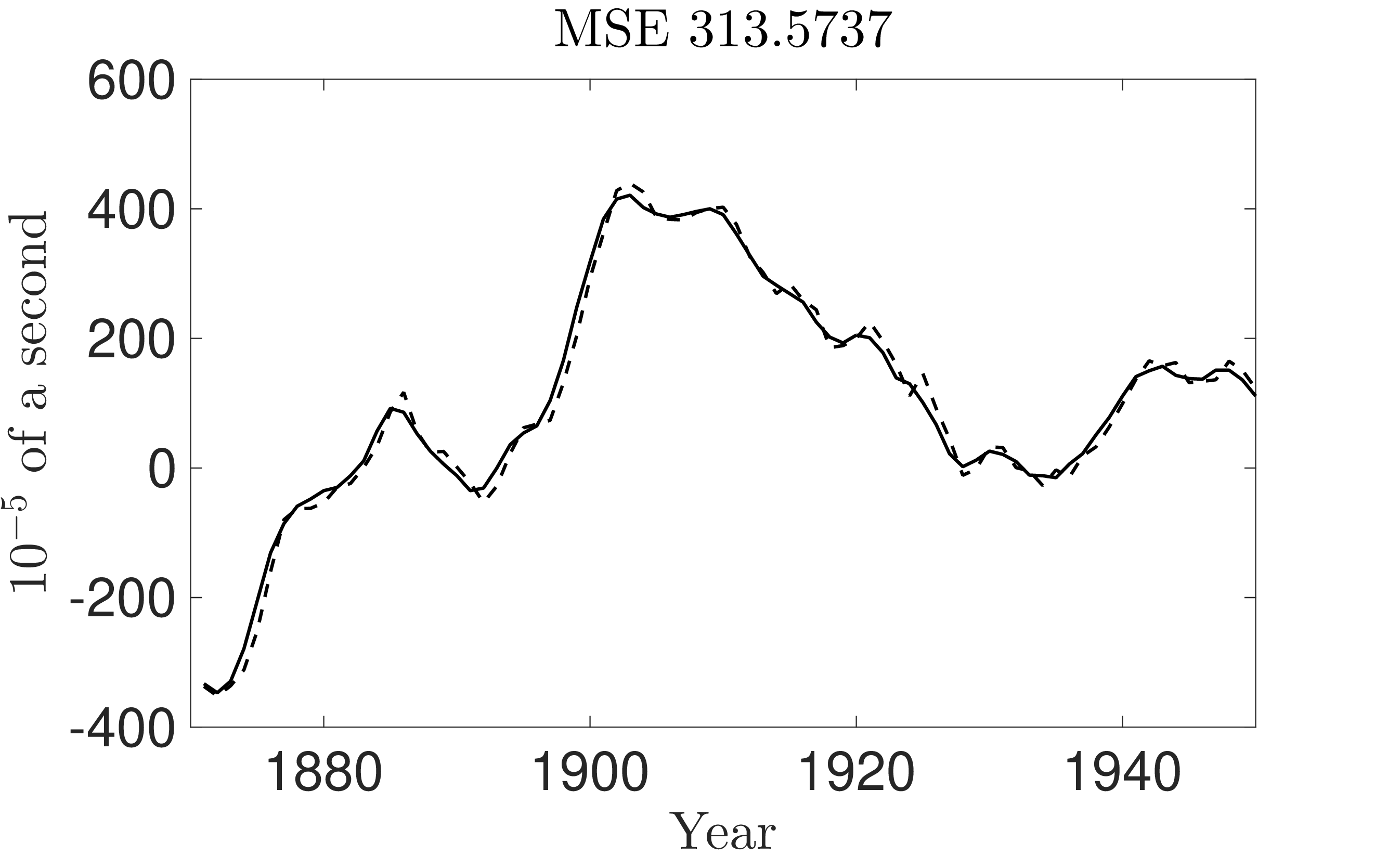}}}
\subfigure[Earthrot using KEM]{\label{fig:pred:earthrot:kear2}
\resizebox{0.47\textwidth}{!}{\includegraphics{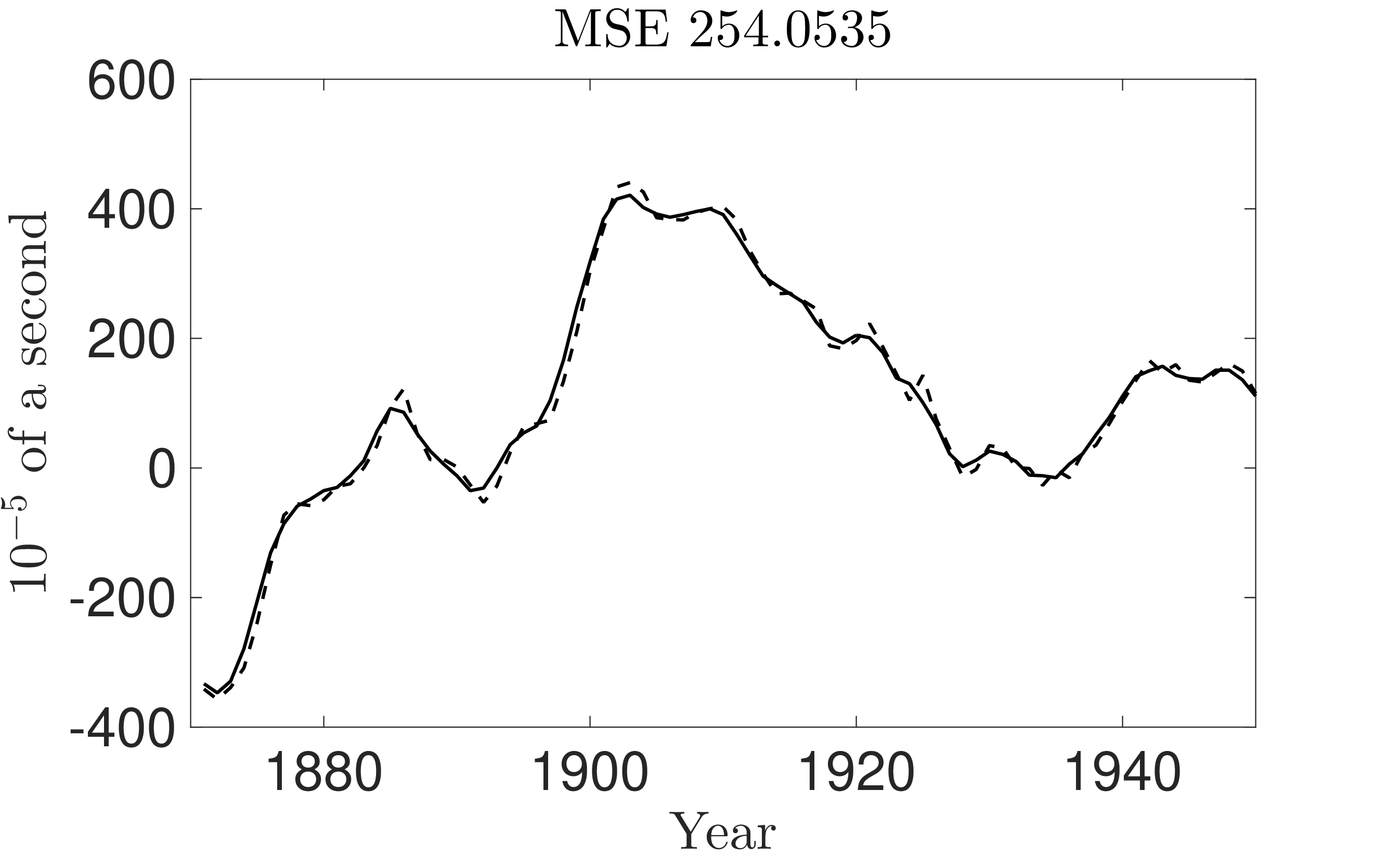}}}
\caption{One-step ahead prediction over the dataset Earthrot given by the linear AR model, the method proposed
  by Kallas et. al. in \cite{Kallas}, and the method based on kernel embeddings
  proposed in this paper. Solid lines are the test data, dashed lines
  are the predictions given by the methods. The title of each figure displays the
  mean squared error between the test data, and the predicted output.}  
\label{fig:dataset:earthrot}
\end{figure}

Figure \ref{fig:dataset:earthrot} shows the one-step ahead prediction
results for the time series Earthrot. For this example, we used
sliding windows of length $51$. The first $50$ observations of each
sliding window were used for training, and the forecast was perfomed
for the time step $51$-st of each sliding window. Since we used the
first $130$ samples from the original time series for the experiment,
and a sliding window of $51$ points, the MSE is computed over a total
of $80$ observation points. We notice that both kernel methods
(figures \ref{fig:pred:earthrot:kallas} and
\ref{fig:pred:earthrot:kear2}) are able to follow the original time
series even from the first time steps, contrary to the linear model
(figure \ref{fig:pred:earthrot:linear}), where the prediction is far
away from the time series. With respect to the $80$ values of $p$ that
were chosen for each method, we computed a simple linear correlation
coefficient between the series of values of $p$'s for the linear
method, and the two kernel approaches. As expected, there is a higher
similarity between the values picked by the KAM, and the KEM, $0.5494$,
compared to $-0.3349$ for the correlation coefficient between the
linear AR model and the KAM, and $0.2405$ for the correlation
coefficient between the linear AR model and KEM.  A further comparison
between the values of $p$ chosen by the kernel methods, show that they
disagreed in $22$ trials out of $80$. With respect to the values of
$\ell$ chosen by the two kernel methods, in only $4$ out of the $80$
trials, both methods chose different bandwidth values. The values for
the MSE show that the method based on kernel embeddings offers the
best performance when compared to the kernel autoregressive method,
and the linear AR model.

\begin{figure}[ht!]
\centering
\subfigure[CO$_2$ using linear AR]{ \label{fig:pred:CO2:linear}
\resizebox{0.47\textwidth}{!}{\includegraphics{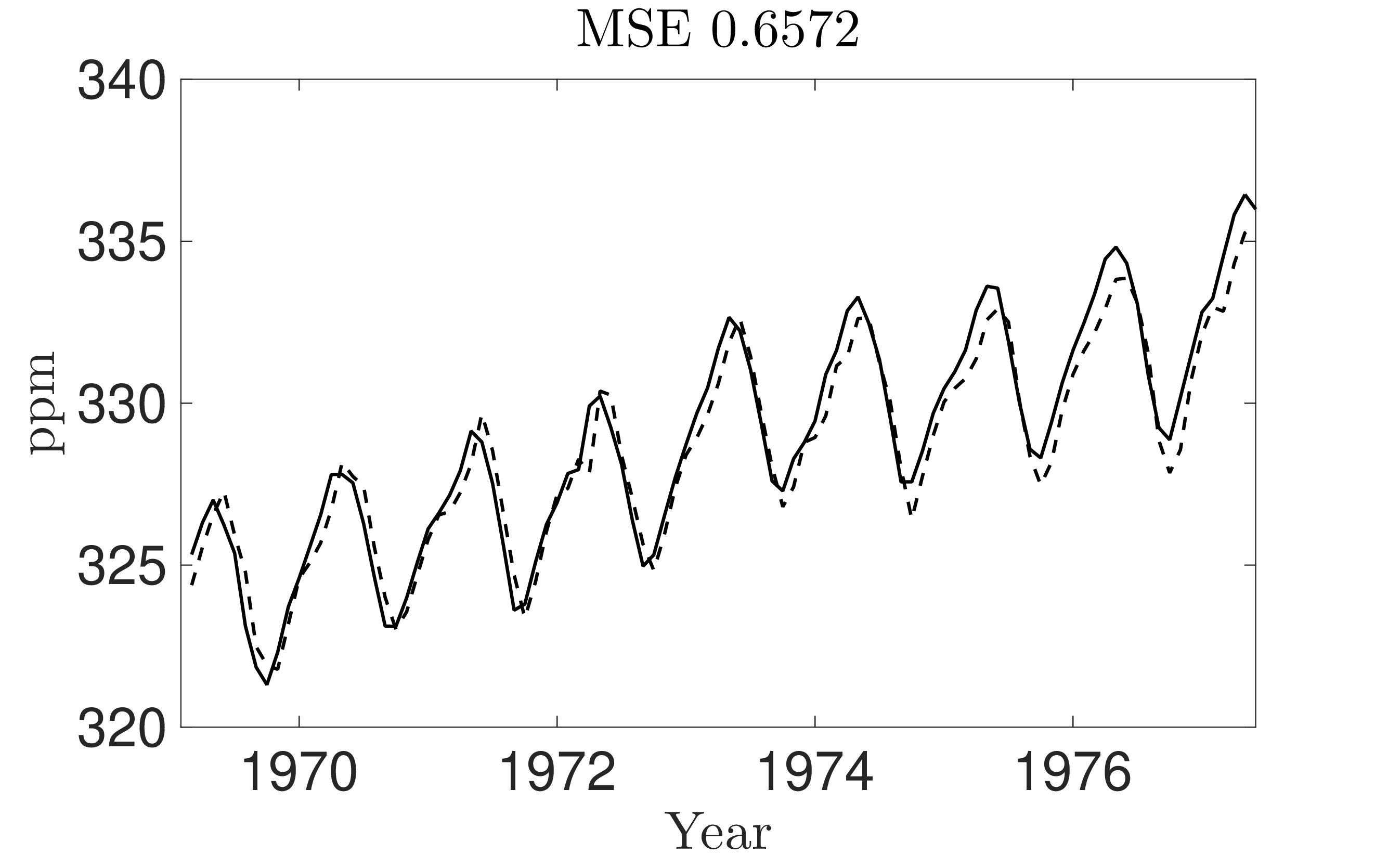}}}
\subfigure[CO$_2$ using KAM]{\label{fig:pred:CO2:kallas}
\resizebox{0.47\textwidth}{!}{\includegraphics{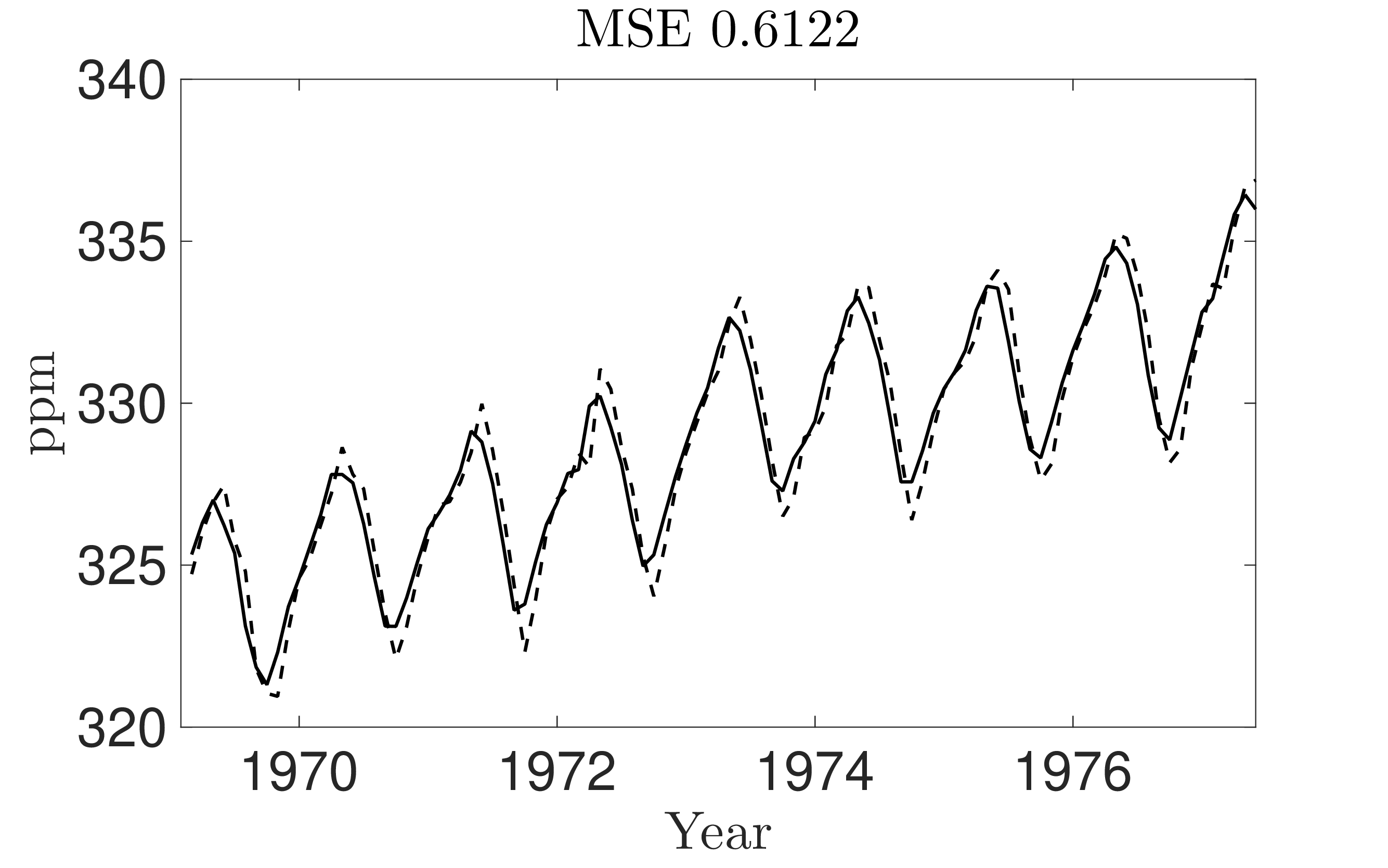}}}
\subfigure[CO$_2$ using KEM]{\label{fig:pred:CO2:kear2}
\resizebox{0.47\textwidth}{!}{\includegraphics{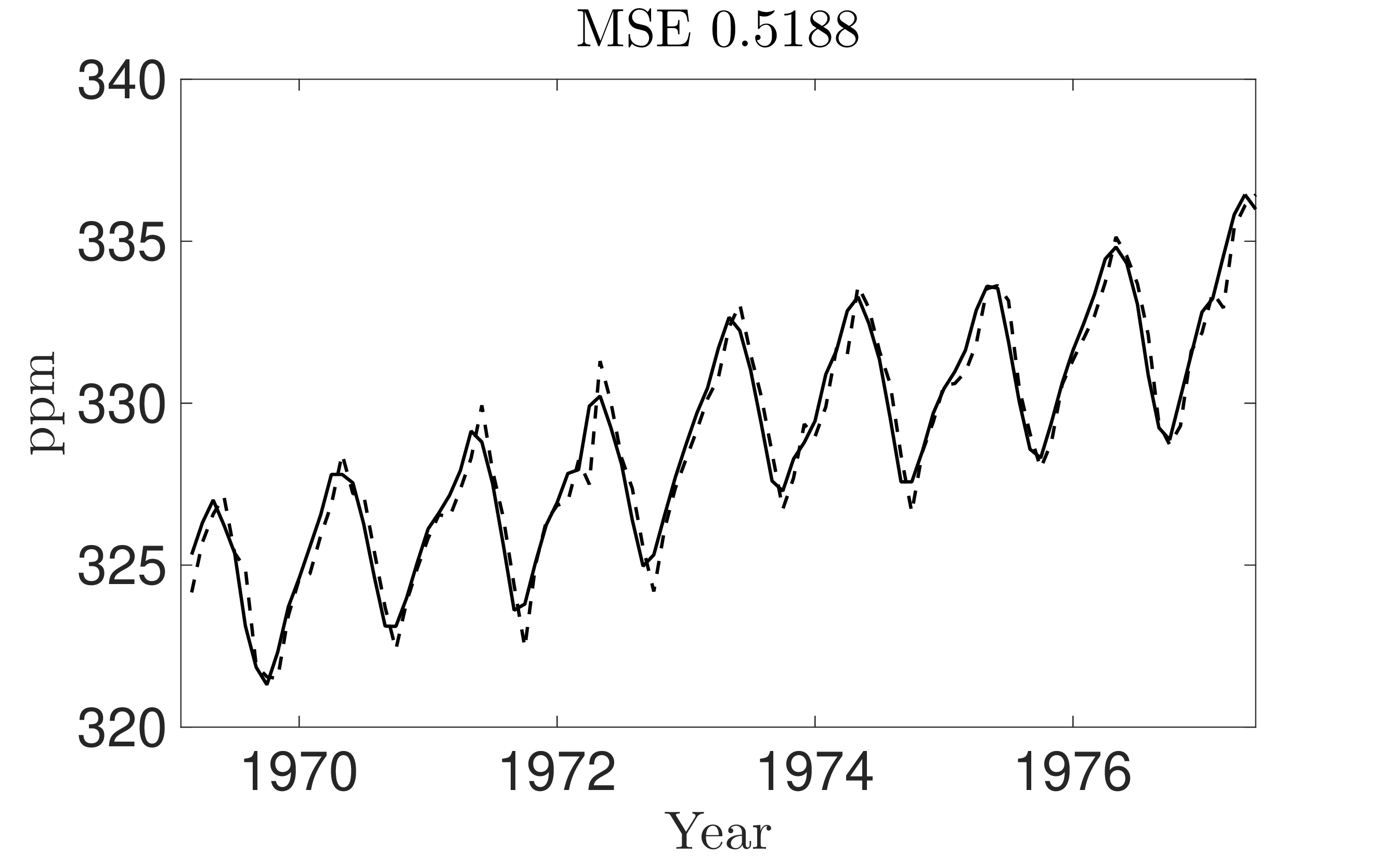}}}
\caption{One-step ahead prediction over the dataset CO$_2$, given by the linear AR model, the method proposed
  by Kallas et. al. in \cite{Kallas}, and the method based on kernel embeddings
  proposed in this paper. Solid lines are the test data, dashed lines
  are the predictions given by the methods. The title of each figure displays the
  mean squared error between the test data, and the predicted output.}  
\label{fig:dataset:co2}
\end{figure}

Figure \ref{fig:dataset:co2} shows the results of one-step ahead
forecasting for the linear AR model, the KAM, and the KEM. As in the
previous example, we used sliding windows of length $51$ samples,
where the first $50$ samples in each window are used for finding
parameters of the models, and the last sample (number $51$) is used to
test the forecasting ability of the methods. From the CO$_2$ time
series that is originally available, we used the first $150$ samples to
assess the prediction performance in several points of the
time-series. Since we use window frames of $51$ points, the MSE error
for the prediction is computed over $100$ samples of the time
series. 

It can be noticed how the KEM method is able to follow more closely
the low and high peak values of the time series, when compared to the
linear method, and KAM. This can be explained by the fact that the
kernel embbeding method is able to take into account the particular
structure in the time series, which for the KAM is lost when averaged. With
respect to the values of $p$, the linear AR model chooses a value of
$p=2$, or $p=5$, mostly. The KAM consistently worked better with $p=2$,
and the KEM with $p=5$. The values chosen for $\ell$ in the kernel
methods were equal $80\%$ of the trails. The MSE values (appearing on
the title of each figure) indicate that the KEM outperforms the linear
AR method and the KAM.

\begin{figure}[ht!]
\centering
\subfigure[MG$_{30}$ using linear AR]{ \label{fig:pred:mg30:linear}
\resizebox{0.47\textwidth}{!}{\includegraphics{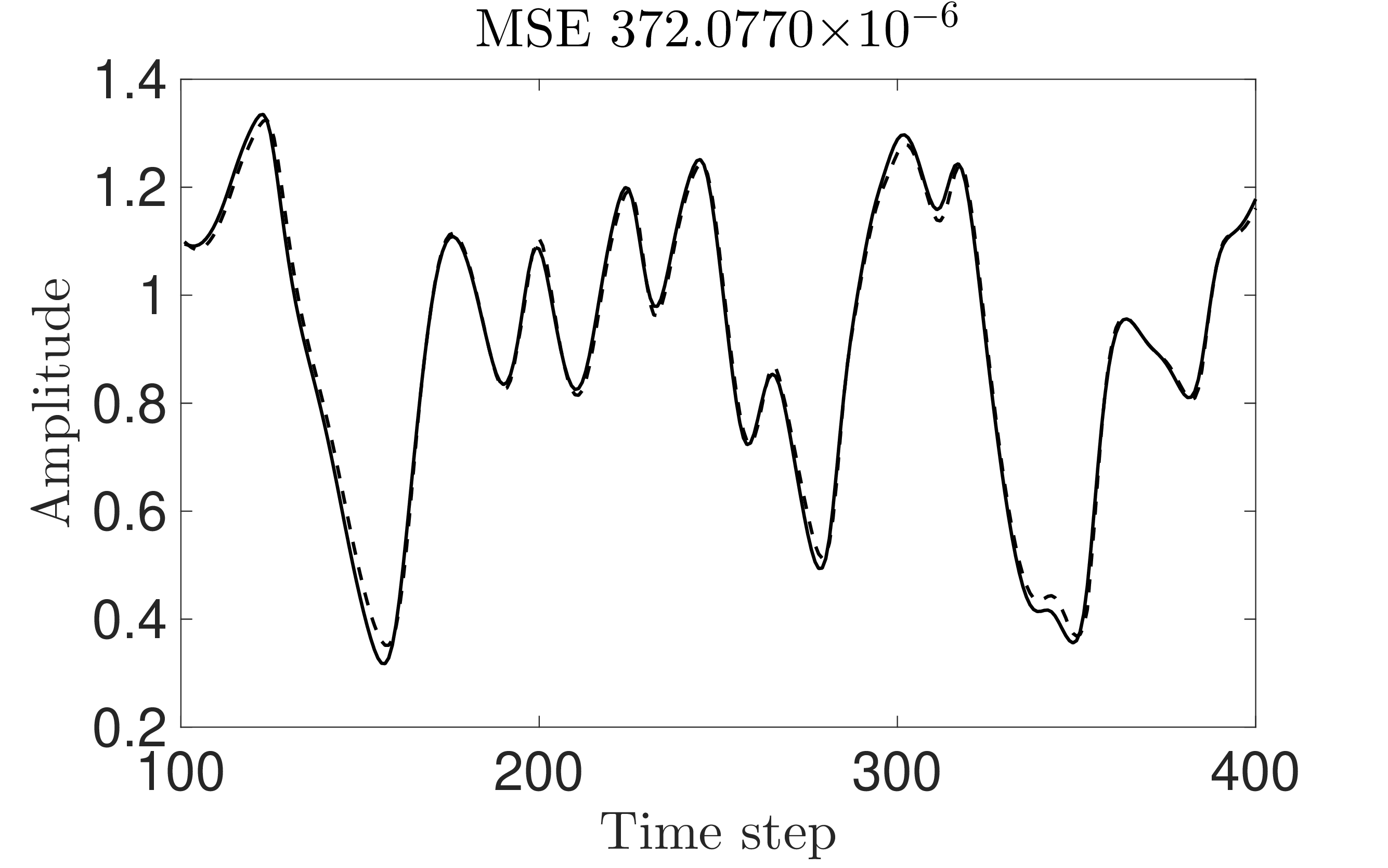}}}
\subfigure[MG$_{30}$ using linear AR]{ \label{fig:pred:mg30:linear:zoom}
\resizebox{0.47\textwidth}{!}{\includegraphics{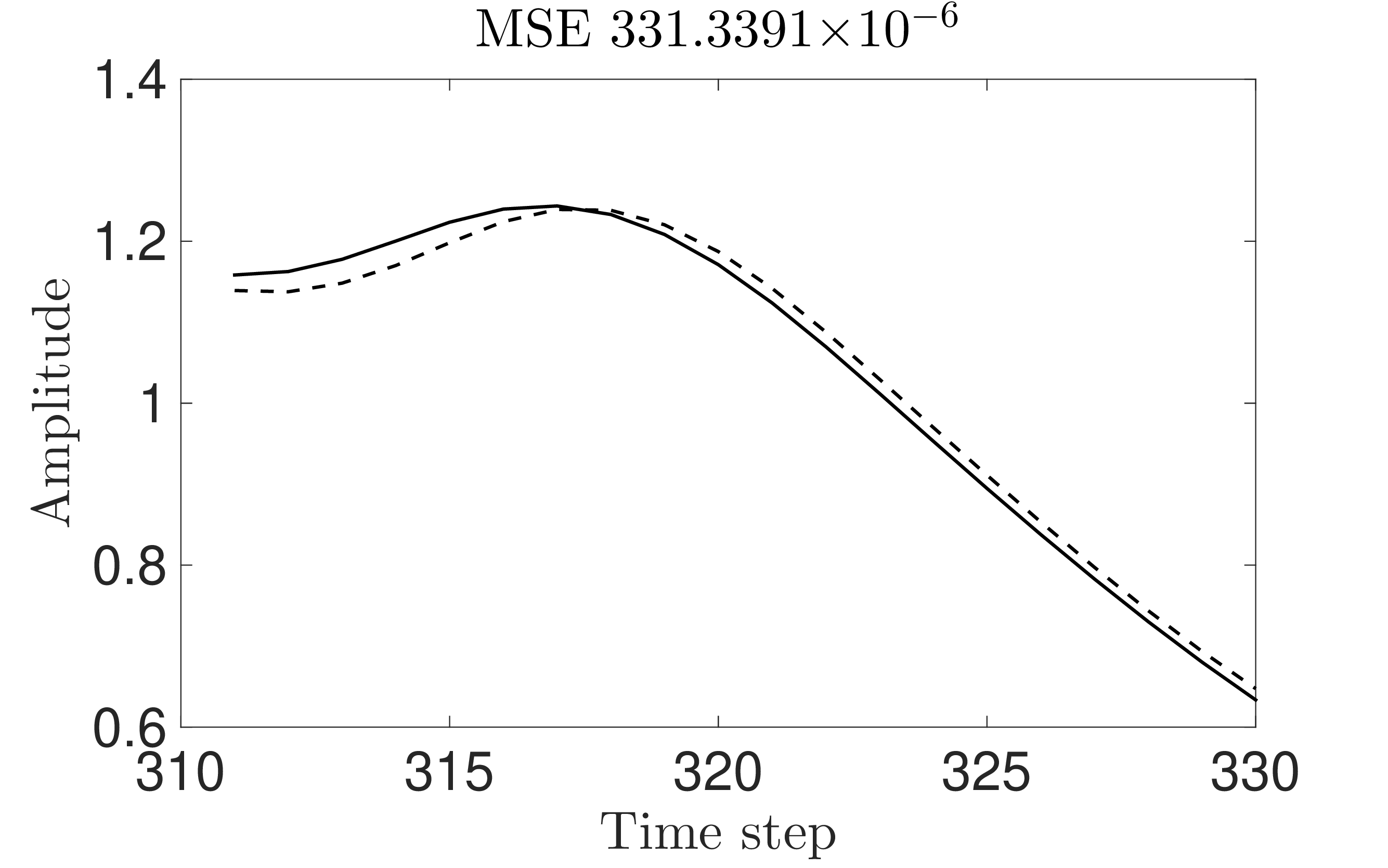}}}
\subfigure[MG$_{30}$ using KAM]{\label{fig:pred:mg30:kallas}
\resizebox{0.47\textwidth}{!}{\includegraphics{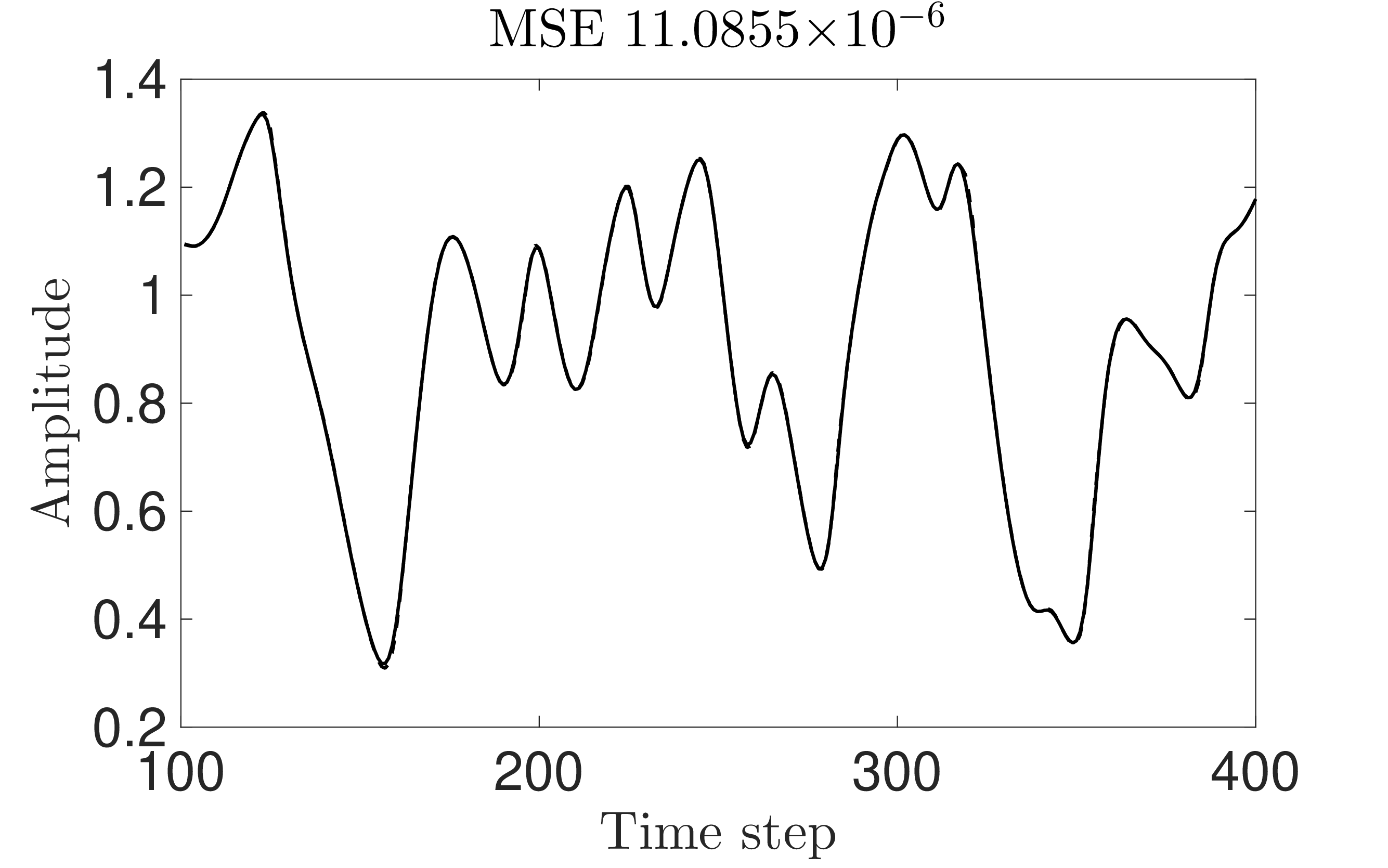}}}
\subfigure[MG$_{30}$ using KAM]{\label{fig:pred:mg30:kallas:zoom}
\resizebox{0.47\textwidth}{!}{\includegraphics{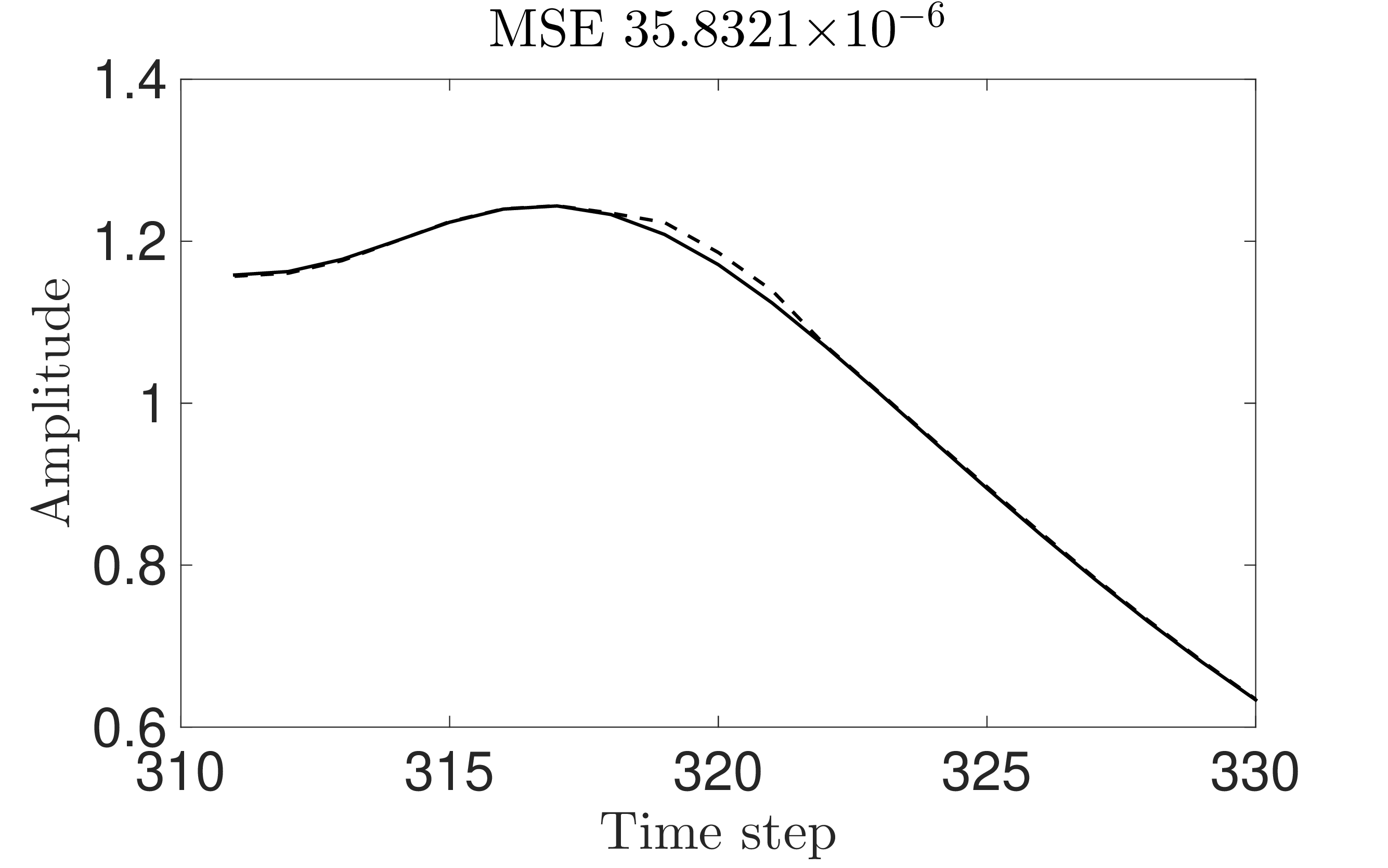}}}
\subfigure[MG$_{30}$ using KEM]{\label{fig:pred:mg30:kear2}
\resizebox{0.47\textwidth}{!}{\includegraphics{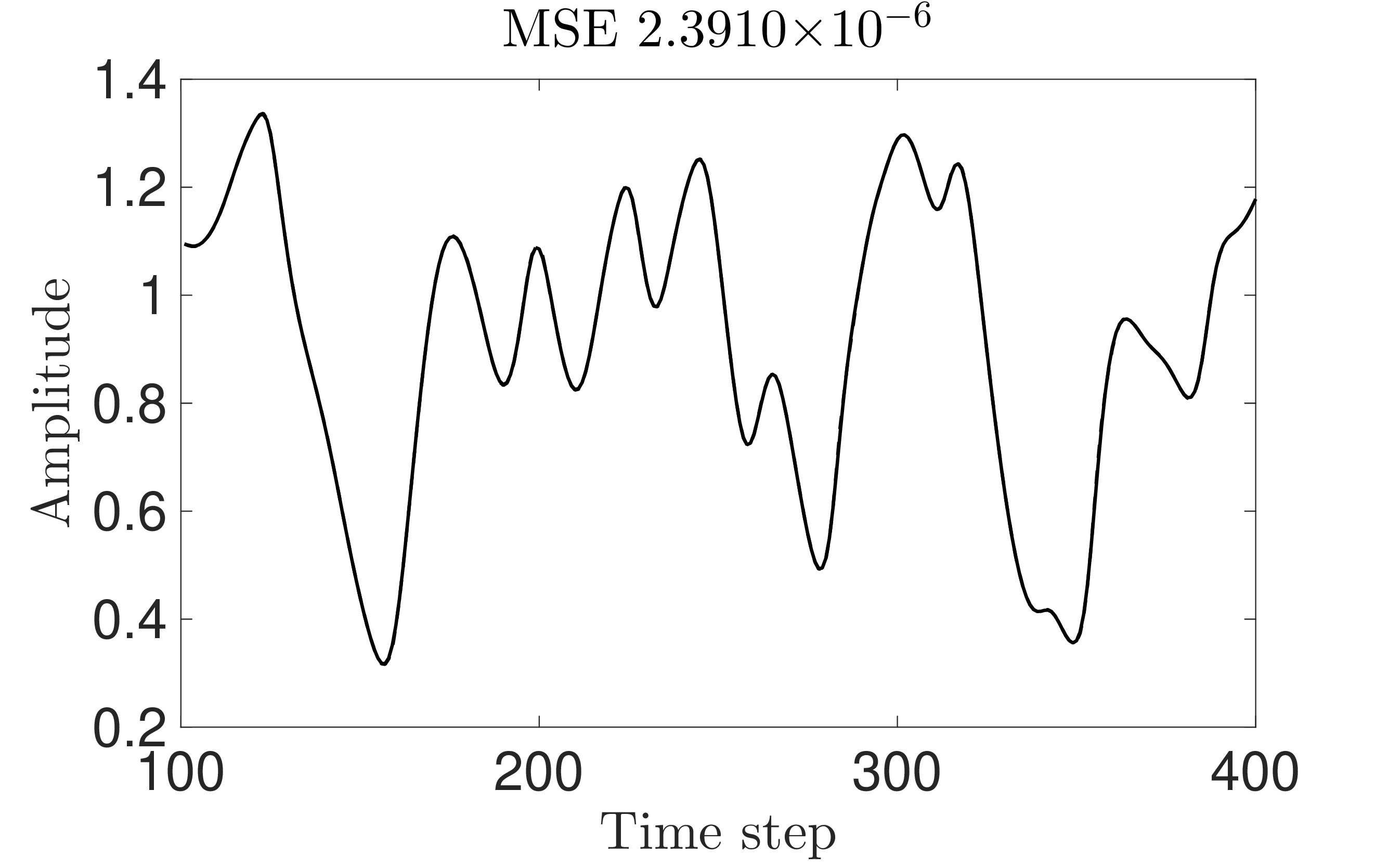}}}
\subfigure[MG$_{30}$ using KEM]{\label{fig:pred:mg30:kear2:zoom}
\resizebox{0.47\textwidth}{!}{\includegraphics{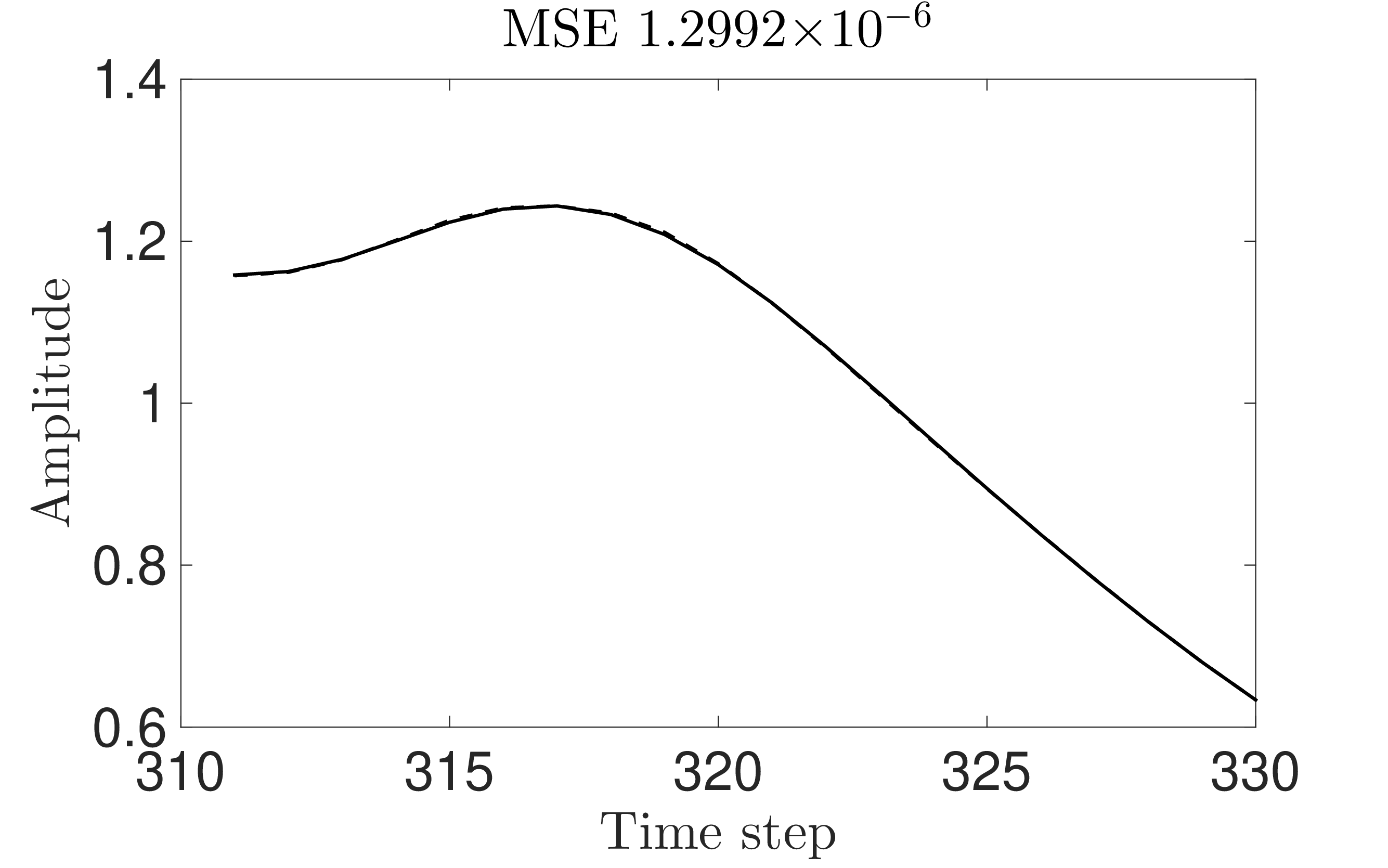}}}
\caption{One-step ahead prediction over the MG$_{30}$ dataset given by
  the linear AR model, the method proposed
  by Kallas et. al.in \cite{Kallas}, and the method based on kernel embeddings
  proposed in this paper. Solid lines are the test data, dashed lines
  are the predictions given by the methods.  Figures
  \ref{fig:pred:mg30:linear}, \ref{fig:pred:mg30:kallas}, and
  \ref{fig:pred:mg30:kear2} show results for the MG$_{30}$
  time-series. Figures \ref{fig:pred:mg30:linear:zoom},
  \ref{fig:pred:mg30:kallas:zoom}, and \ref{fig:pred:mg30:kear2:zoom}
  show results for the MG$_{30}$ 
time-series within a shorter time period, between time steps 311 and
330. The title of each figure displays the
  mean squared error between the test data, and the predicted output.}  
\label{fig:dataset:mg30}
\end{figure}

Figure \ref{fig:dataset:mg30} shows the one-step ahead prediction for
the Mackey-Glass chaotic time series. For this time series, we use
sliding windows of length $101$. The first $100$ samples of the
sliding window are used for
training the models, and the sample $101$-st is used for one-step ahead
prediction. We perform the one-step ahead prediction over consecutive
$300$ samples, one at a time, and the MSE reported is the average over
these $300$ one-step ahead forecasting values.  It can be noticed from
figures \ref{fig:pred:mg30:linear}, \ref{fig:pred:mg30:kallas}, and
\ref{fig:pred:mg30:kear2} that the methods based on kernels yield
better prediction results than the linear method. Since it seems that
qualitatively, the prediction performance for KAM and KEM is similar,
we included additional figures where we zoom in a particular range
where the difference in performace can be noticed. Figures
\ref{fig:pred:mg30:linear:zoom}, \ref{fig:pred:mg30:kallas:zoom}, and
\ref{fig:pred:mg30:kear2:zoom} show results for the MG$_{30}$
time-series within a shorter time period, between time steps 311 and
330. With respect to the $p$ values, the linear model favored
a value of $p=5$ (250 over the 300 trials). The KAM and the KEM predominantly used
higher values of $p$: 70 for $p=4$, and 163 for $p=5$, for the KAM;
and 43 for $p=4$, and 257 for $p=5$, for the KEM.  In contrast to the
experiments above, this time the kernel methods only selected the same
value for $\ell$ in 76 cases out of 300. In the terms of the average
MSE over the 300 trials, the experiment shows that both kernel mehods
outperform the linear AR method. The MSE obtained by the KEM is lower
than the one obtained by KAM.

\begin{figure}[ht!]
\centering
\subfigure[Lorenz using linear AR]{ \label{fig:pred:lorenz:linear}
\resizebox{0.47\textwidth}{!}{\includegraphics{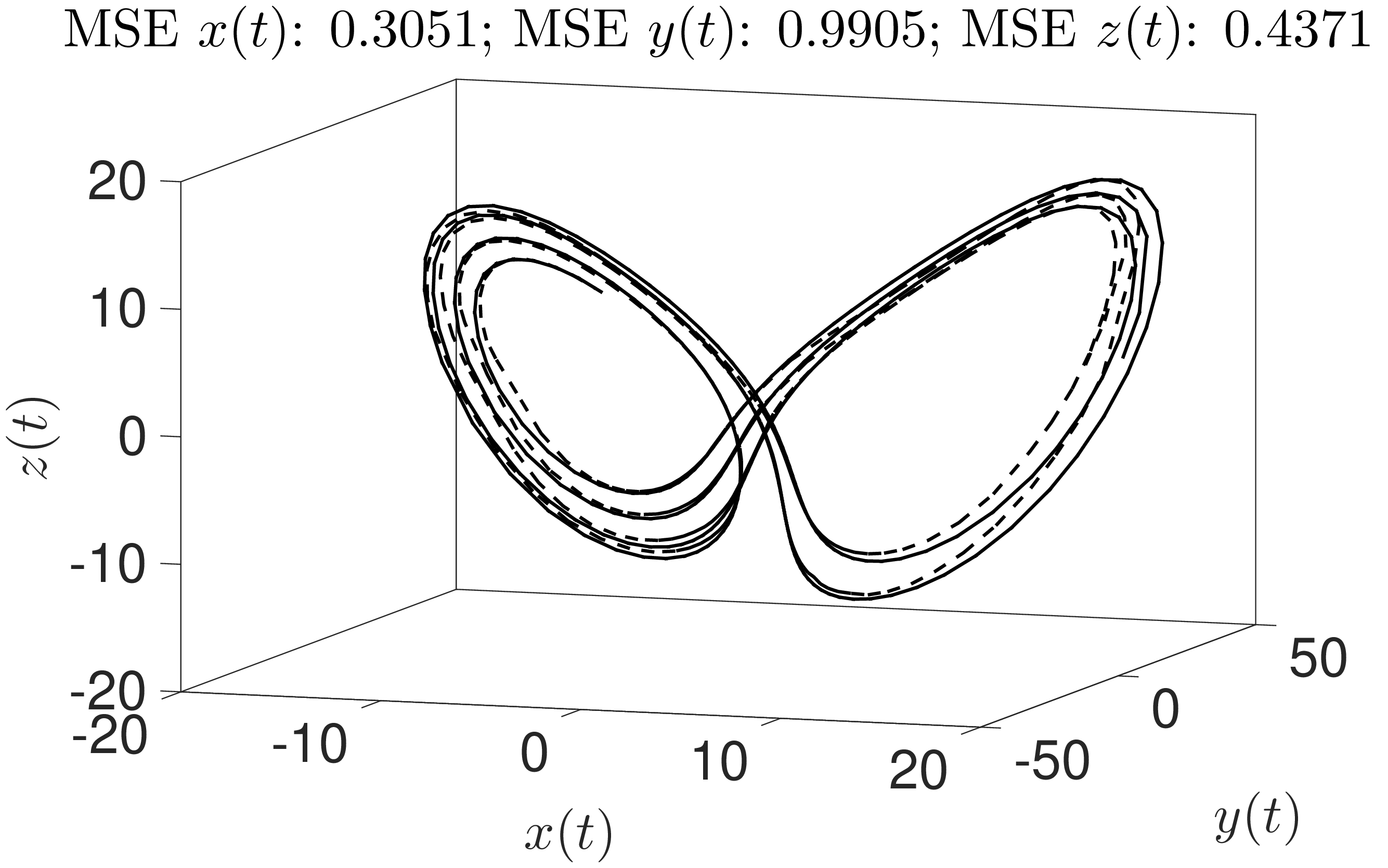}}}
\subfigure[Lorenz using KAM]{\label{fig:pred:lorenz:kallas}
\resizebox{0.47\textwidth}{!}{\includegraphics{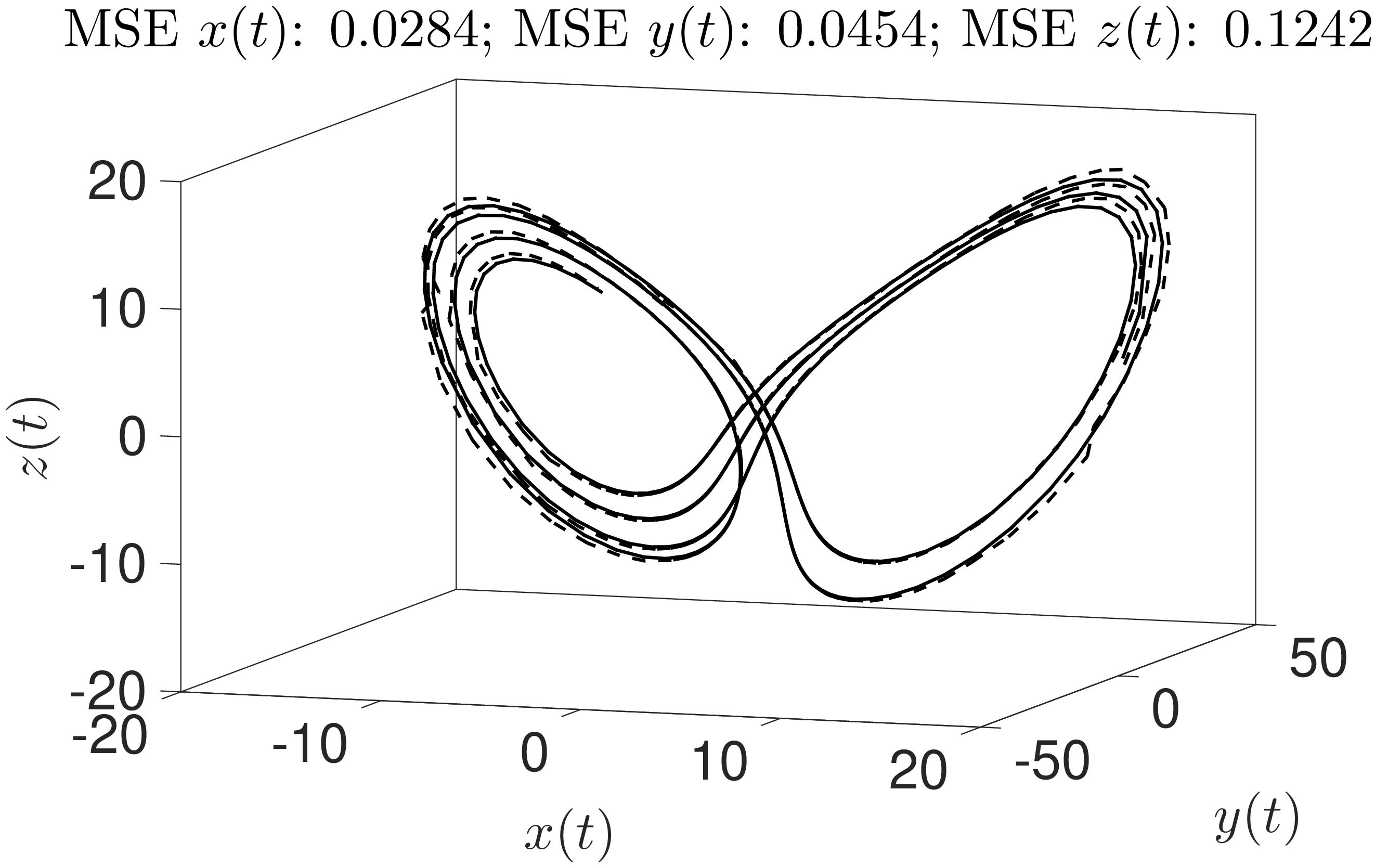}}}
\subfigure[Lorenz using KEM]{\label{fig:pred:lorenz:kear2}
\resizebox{0.47\textwidth}{!}{\includegraphics{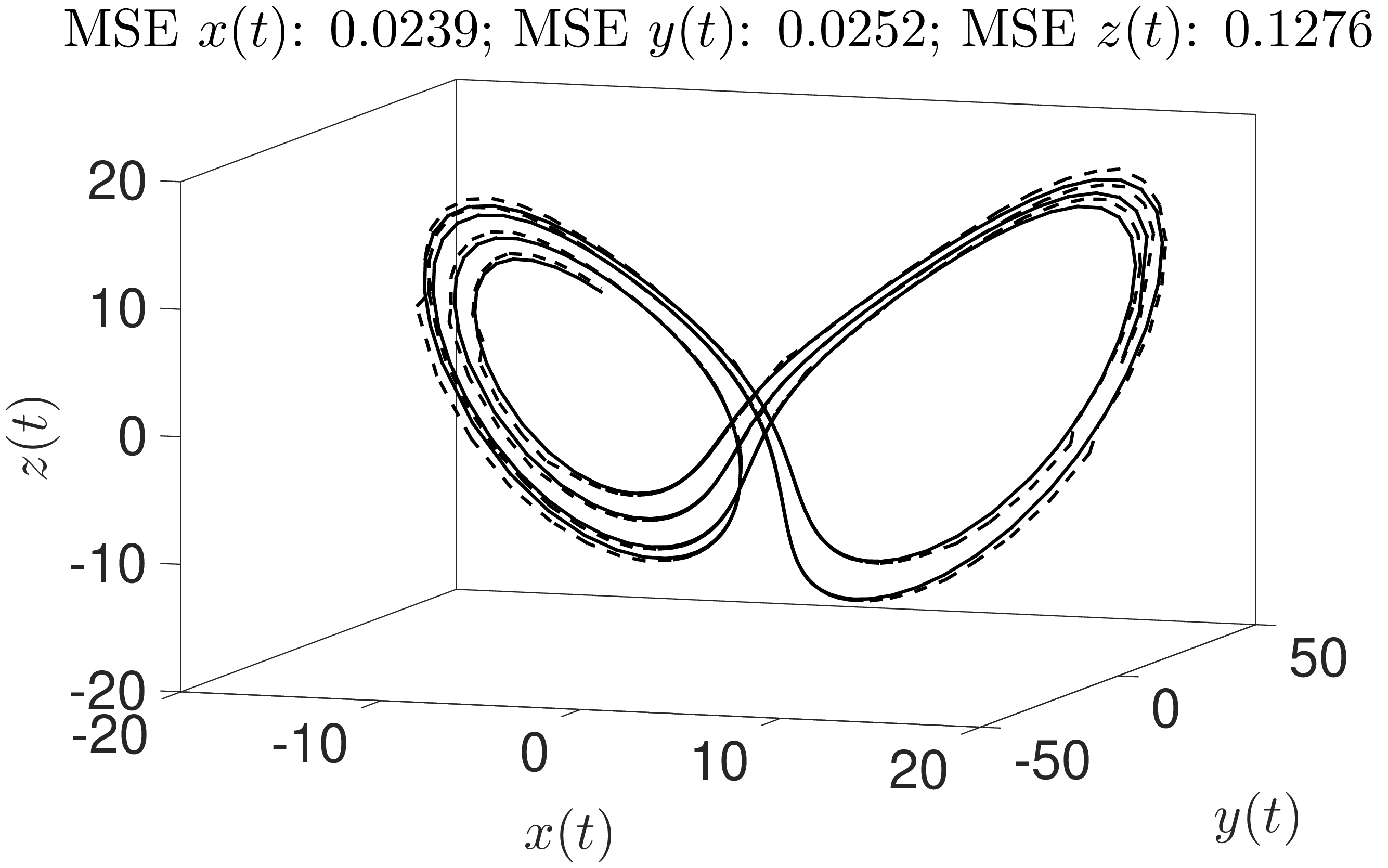}}}
\caption{One-step ahead prediction over the Lorenz dataset given by
  the linear AR model, the method proposed
  by Kallas et. al. (2013), and the method based on kernel embeddings
  proposed in this paper. Solid lines are the test data, dashed lines
  are the predictions given by the methods. The title of each figure displays the
  mean squared error between the test data, and the predicted output.}  
\label{fig:dataset:lorenz}
\end{figure}

Figure \ref{fig:dataset:lorenz} shows the prediction results over the
Lorenz dataset. As explained before, prediction is performed over each
component $(x(t), y(t), z(t))$ of the 3D time series, in an
independent manner. For each of the three time series, we use sliding
windows of length $101$, where the forecasting is done over the last
time step of each frame. The prediction performance is evaluated over
$300$ successive frames, all of them of length $101$. With respect to
the order $p$ for the different models, the linear AR model picked
$p=2$ almost $22\%$ of the time, and $p=5$ almost $75\%$ of the
time. The KAM chose $p=2$ almost $70\%$ of all trials, and $p=3$
almost $22\%$ of all the repetitions. Finally, the KEM picked $p=2$ almost
$42\%$ of the time, $p=4$ almost $23\%$ of the time, and $p=5$
approximately $30\%$ of the trials. As for the previous experiments,
the kernel methods outperform the linear AR model. Although the
prediction error of the KAM for $z(t)$ is lower than the prediction
error for the KEM, on average, the KEM outperforms the KAM.

\begin{table}[ht!]
\centering
\caption{Mean squared error for the test data and the predicted
  outputs, given by the linear autoregressive process (Linear AR), the kernel
  autoregressive model proposed by Kallas et al in \cite{Kallas} (KAM), and the kernel
  embeddings of autoregressive processes proposed in this paper
  (KEM). The values of the MSE for the MG$_{30}$ should be multiplied
  by $10^{-6}$. }\label{table:pred:complete}
\begin{tabular}{ c  c  c  c}
\hline
 Database & Linear AR & KAM & KEM  \\
\hline
Earthrot & $689.3491$  & $313.5737$ & $\mathbf{254.0535}$  \\ 
CO$_2$ & $0.6572$  & $0.6122$ & $\mathbf{0.5188}$  \\ 
MG$_{30}$ & $372.0770$  & $11.0855 $ & $\mathbf{2.3910}$  \\ 
Lorenz $x(t)$ & $0.3051$  & $0.0284$ & $\mathbf{0.0239}$  \\ 
Lorenz $y(t)$ & $0.9905$  & $0.0454$ & $\mathbf{0.0252}$  \\ 
Lorenz $z(t)$ & $0.4371$  & $\mathbf{0.1242}$ & $0.1276$  \\ 
\hline
\end{tabular}
\end{table}

Table \ref{table:pred:complete} shows a summary of the MSE obtained by the
linear AR model, the kernel autoregressive model, and the kernel
embedding method for the four datasets. It is clear from that table
that the method that uses the kernel embeddings lead to better
results, except of the component $z(t)$ of the Lorenz time series.

\begin{table}[ht!]
\centering
\caption{Mean squared error for the test data and the predicted
  outputs, given by a neural network (NN), a Gaussian process
  regressor (GP), the kernel
  autoregressive model proposed by Kallas et al in \cite{Kallas} (KAM), and the kernel
  embeddings of autoregressive processes proposed in this paper
  (KEM). The values of the MSE for the MG$_{30}$ should be multiplied
  by $10^{-6}$.}\label{table:pred:allmethods}
\begin{tabular}{ c  c  c  c  c}
\hline
 Database & NN & GP & KAM & KEM  \\
\hline
Earthrot & $827.8469$  & $570.1689$ & $182.3038$ & $\mathbf{134.2564}$ \\ 
CO$_2$ & $0.6027$  & $0.4631$ & $0.4107$ & $\mathbf{0.3177}$ \\ 
MG$_{30}$ & $41.4519$  & $2.0991$ & $6.3064$ & $\mathbf{1.1052}$ \\ 
Lorenz $x(t)$ & $0.0595$  & $0.0118$ & $0.0088$ & $\mathbf{0.0037}$ \\ 
Lorenz $y(t)$ & $0.1114$  & $0.0129$ & $0.0146$ & $\mathbf{0.0038}$ \\ 
Lorenz $z(t)$ & $0.2041$  & $0.0263$ & $0.0211$ & $\mathbf{0.0140}$ \\ 
\hline
\end{tabular}
\end{table}

Table \ref{table:pred:allmethods} shows the performance of neural
networks, Gaussian processes, and the kernel autoregressive method
compared to the performance of the kernel embeddings proposed in this
paper. The value of $w$ for all the time series was fixed to $50$, and
the one-step ahead forecasting was performed for $80$ time steps for
Earthrot, $100$ time steps for CO$_2$, and $300$ for both MG$_{30}$,
and Lorenz. The numerical optimization methods used for NN and GP are
based on gradient-descent-like procedures, which heavily depend on a
good parameter initialization to deliver sensible results. A bad
parameter initialization often leads to poor prediction
performance. In order to reduce the number of outliers for the
prediction for NN and GP, we only computed the mean for those squared
errors that were between quartiles $25$-th and $75$-th of all the
squared errors computed for each time series. For a fair comparison,
we also computed the MSE for the KAM, and the MSE for the KEM removing
outliers, as explained before. We noticed from table
\ref{table:pred:allmethods} that the methods based on kernels, GP, KAM
and KEM, yield better prediction performance than NN. Gaussian
processes outperform KAM for the MG$_{30}$ time series, and the
component $y(t)$ of the Lorenz time series. The KEM method shows
improved performance over all the other competing models.

\section{Conclusions}
\label{sec:conclusions}

In this paper, we have introduced kernel embeddings of joint probability distributions by means of an autoregressive process of order $p$ placed over covariance operators. The solution to the model is done through a Yule-Walker system of equations for empirical estimates of the cross-covariance operators. Predictions in the input space are performed by solving a pre-image problem, for which a fixed-point algorithm is developed. Experimental results show that the method proposed here outperforms several non-linear versions of the autoregressive model, in the task of one-step ahead forecasting of time series. An important extension of this line work would be the formulation of a non-linear vector-valued autoregressive model, for which coefficients $\{\alpha_j\}_{j=1}^p$ would need to be considered as more general linear operators.

\section*{Acknowledgements} 

E. A. Valencia is being partly funded by University Tecnol\'ogica de
Pereira. The authors would like to thank Colciencias and British
Council for funding under the project ``Hilbert space embeddings of
Autoregressive processes''. The authors would also like to thank
Arthur Gretton, Zolt\'an Szab\'o and Kenji Fukumizu for their insightful
comments and suggestions.

\bibliographystyle{plain}
\bibliography{kaemEdgar}

\end{document}